\documentclass[preprint,review,12pt]{elsarticle}
\usepackage{url}
\usepackage{xcolor}
\usepackage{times}
\usepackage{epsfig}
\usepackage{graphicx}
\usepackage{amsmath}
\usepackage{color}
\usepackage{colortbl}
\usepackage{xcolor}
\usepackage{csquotes}
\usepackage{tipa}
\usepackage{upgreek}

\newcommand{\argmax}{\mathop{\mathrm{argmax}}} 

\usepackage[pagebackref=true,breaklinks=true,colorlinks,bookmarks=false]{hyperref}

\usepackage[labelfont=bf,skip=0pt]{caption}

\usepackage{xspace}
\newcommand*{\eg}{e.g.\@\xspace}
\newcommand*{\etal}{\textit{et al}.\@\xspace}
\newcommand*{\ie}{i.e.\@\xspace}

\makeatletter
\newcommand*{\etc}{%
    \@ifnextchar{.}%
        {etc}%
        {etc.\@\xspace}%
}

\makeatletter
\def\ps@pprintTitle{%
   \let\@oddhead\@empty
   \let\@evenhead\@empty
   \let\@oddfoot\@empty
   \let\@evenfoot\@oddfoot
}
\makeatother
\definecolor{newcolor}{rgb}{.8,.349,.1}

\begin{document}

\begin{frontmatter}

	\title{Deep Ancient Roman Republican Coin Classification via Feature Fusion and Attention}

	\author[1,2]{Hafeez Anwar\corref{cor1}\fnref{label2}} 
	\cortext[cor1]{Corresponding author:}
	\ead{hafeez.anwar@fau.de}
	\author[3,4]{Saeed Anwar\fnref{label2}}
	\author[5]{Sebastian Zambanini}
	\author[4]{Fatih Porikli}
	
	\address[1]{Department of Electrical and Computer Engineering, COMSATS University Islamabad, Attock Campus Pakistan}
	\address[2]{Interdisciplinary Center for Digital Humanities and Social Sciences, Friedrich-Alexander University (FAU), Germany}
	\address[3]{Data61-CSIRO, Australia.}
	\address[4]{College of Engineering and Computer Science, The Australian National University, Australia.}
	\address[5]{Computer Vision Lab, TU Wien, Austria}
	\fntext[label2]{Shows equal contribution}
	
	\begin{abstract}
		We perform the classification of ancient Roman Republican coins via recognizing their reverse motifs where various objects, faces, scenes, animals, and buildings are minted along with legends. Most of these coins are eroded due to their age and varying degrees of preservation, thereby affecting their informative attributes for visual recognition. Changes in the positions of principal symbols on the reverse motifs also cause huge variations among the coin types. Lastly, in-plane orientations, uneven illumination, and a moderate background clutter further make the classification task non-trivial and challenging. 
		
		To this end, we present a novel network model, CoinNet, that employs compact bilinear pooling, residual groups, and feature attention layers. Furthermore, we gathered the largest and most diverse image dataset of the Roman Republican coins that contains more than 18,000 images belonging to 228 different reverse motifs. On this dataset, our model achieves a classification accuracy of more than \textbf{98\%} and outperforms the conventional bag-of-visual-words based approaches and more recent state-of-the-art deep learning methods. We also provide a detailed ablation study of our network and its generalization capability.   
		
	\end{abstract}
	
	\begin{keyword}
		Coins dataset, Compact bilinear pooling, Convolutional networks, Visual attention, Residual blocks, Deep learning in art's history, Roman Republican coins.
	\end{keyword}
	
\end{frontmatter}

\section{Introduction}

Coins have been the dominant type of currency in human history, and for this reason, their recognition is of significant interest in both the academic and economic worlds. In archaeology, history, and art, ancient coins reveal an enriched understanding of cultural and historical events. In commerce, they are valuable trading items due to their antiquity. In contrast to present-day coins, recognizing and understanding ancient coins requires in-depth and highly specialized domain expertise. This challenge is partly attributed to severe abrasions due to their age, yet the main complexity stems from their finely granulated class structure. For instance, the Roman Republican coins have over 1900 classes and subclasses defined in standard reference books~\cite{crawford1974roman}. With such a large number of categories, the ancient coins further face an additional challenge of \enquote{rarity} where the worldwide count of specimen for some classes is considerably low. Consequently, there is a clear interest in automatically extracting information about an unknown coin, and several works in the past \cite{anwar2013bag,anwar2015ancient,arandjelovic2010automatic} have attempted to address this problem. As being a visual recognition task, recent state-of-the-art convolutional neural network (CNN) based models~\cite{cooper2019understanding,kim2016discovering,schlag2017ancient} have also been applied, albeit their strong dependency on comprehensive and annotated image datasets.

\begin{figure}[t]
\begin{center}
\resizebox{0.7\columnwidth}{!}{
\begin{tabular}{c@{ }c@{ }c@{ }c}

     \includegraphics[width=0.33\linewidth, keepaspectratio]{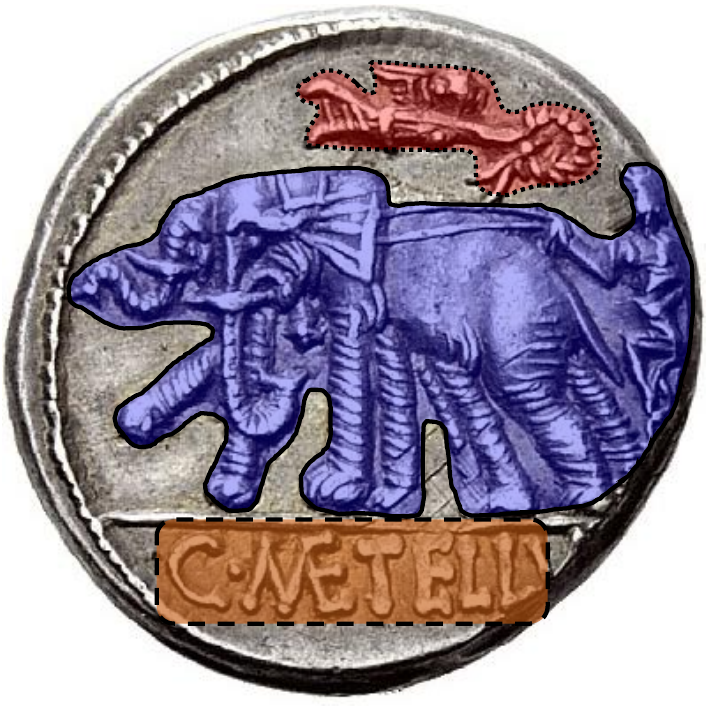}&  
    \includegraphics[width=0.33\linewidth, keepaspectratio]{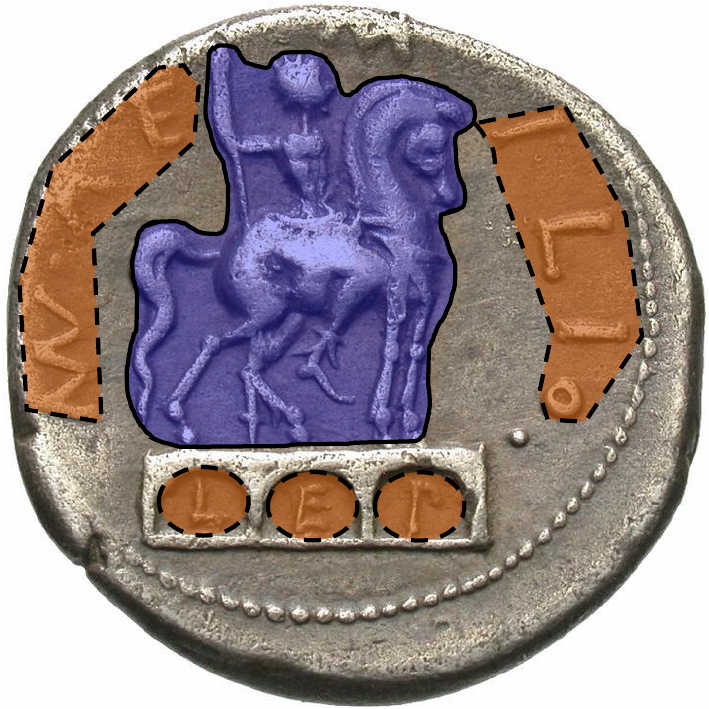}&  
    \includegraphics[width=0.33\linewidth, keepaspectratio]{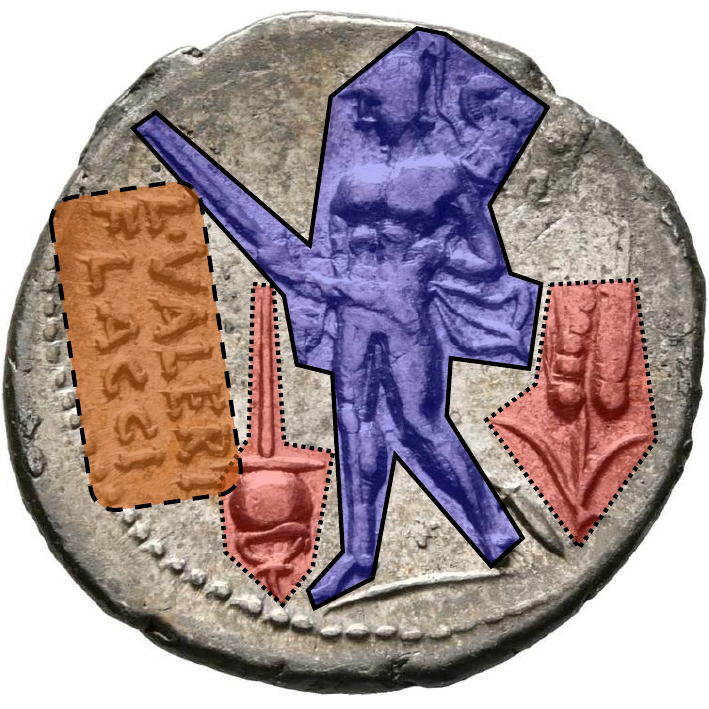}&  
    \includegraphics[width=0.33\linewidth, keepaspectratio]{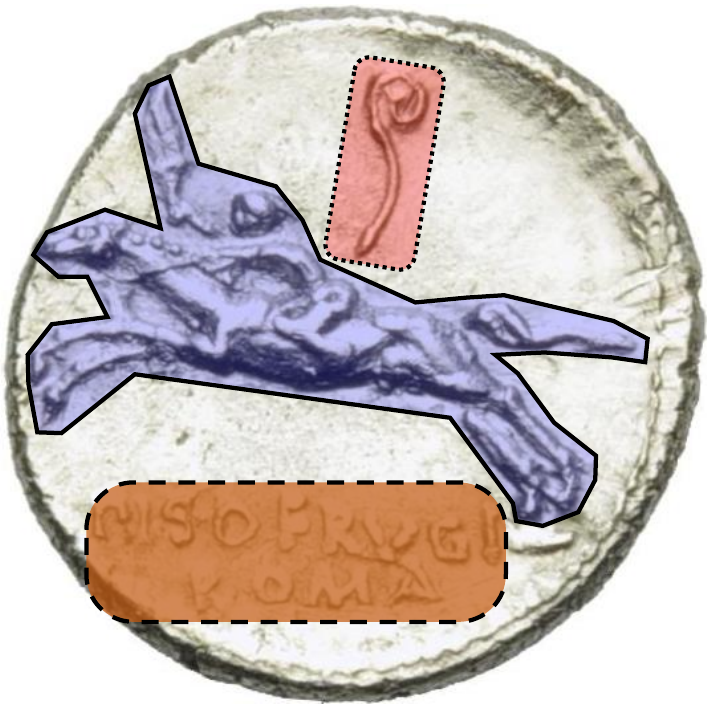}\\  
    \includegraphics[width=0.33\linewidth, keepaspectratio]{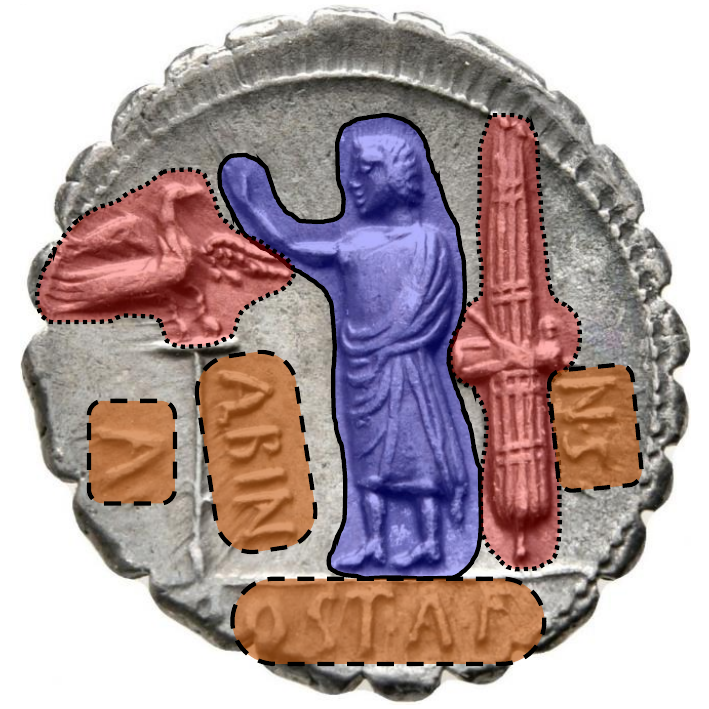}&  
    \includegraphics[width=0.33\linewidth, keepaspectratio]{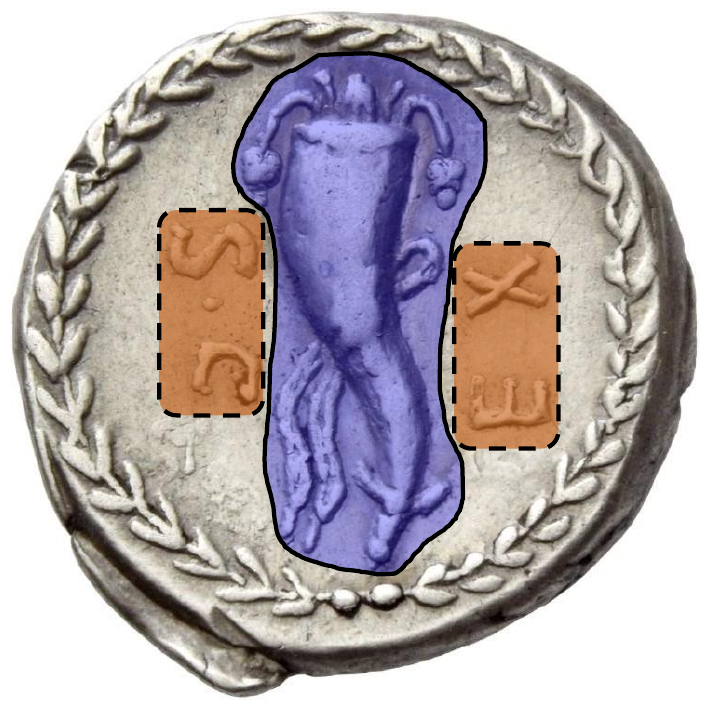}&  
    \includegraphics[width=0.33\linewidth, keepaspectratio]{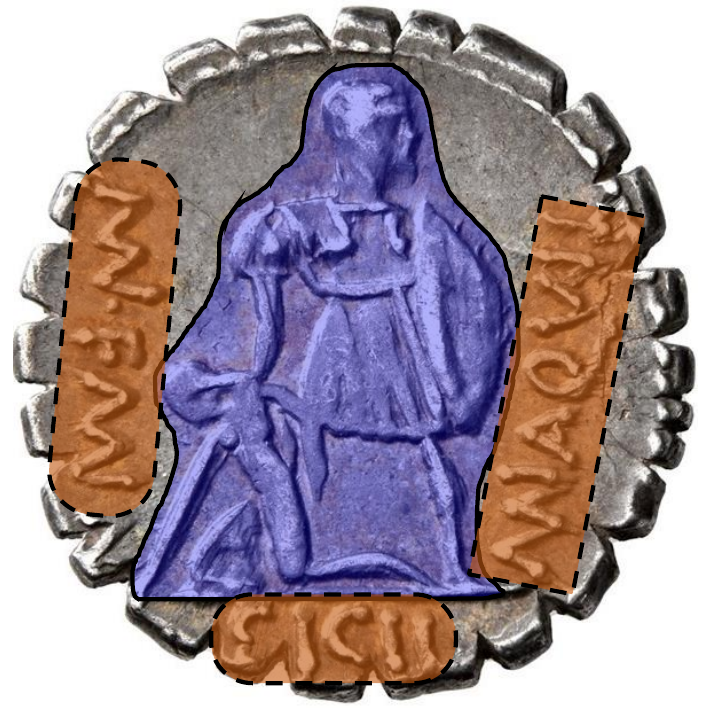}&  
    \includegraphics[width=0.33\linewidth, keepaspectratio]{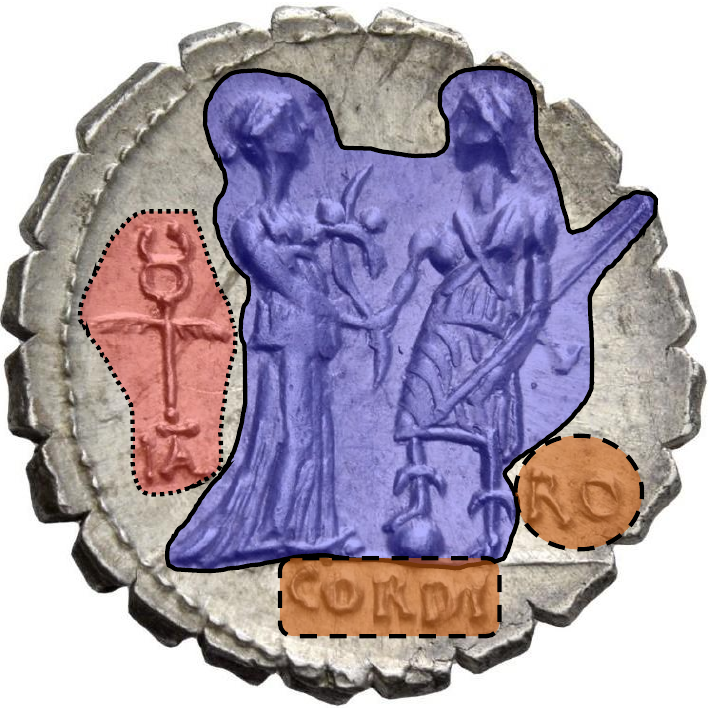}\\  
    \includegraphics[width=0.33\linewidth, keepaspectratio]{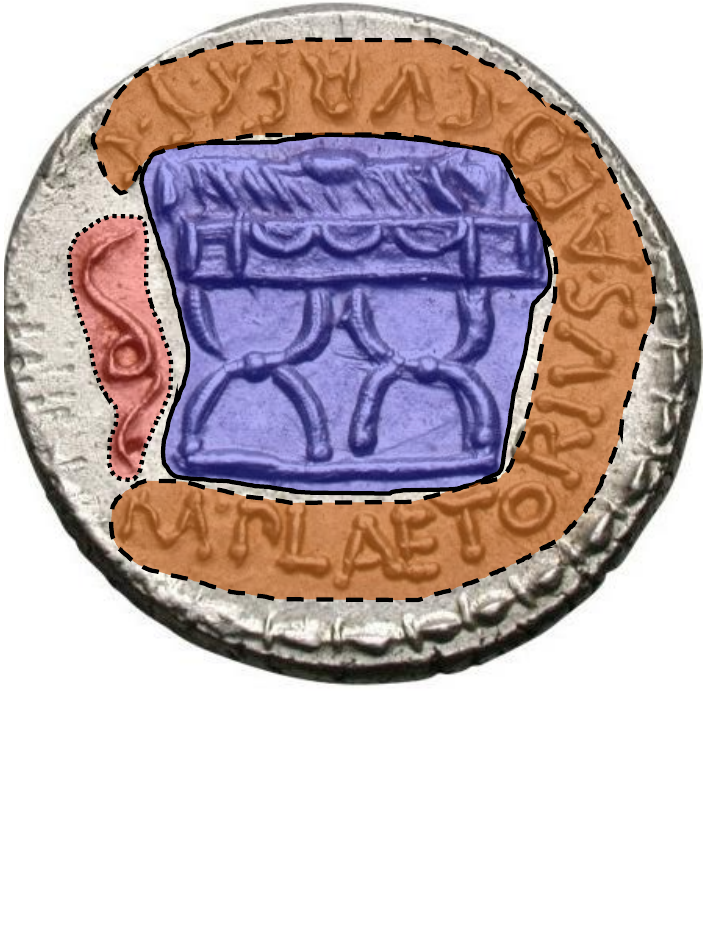}&  
    \includegraphics[width=0.33\linewidth, keepaspectratio]{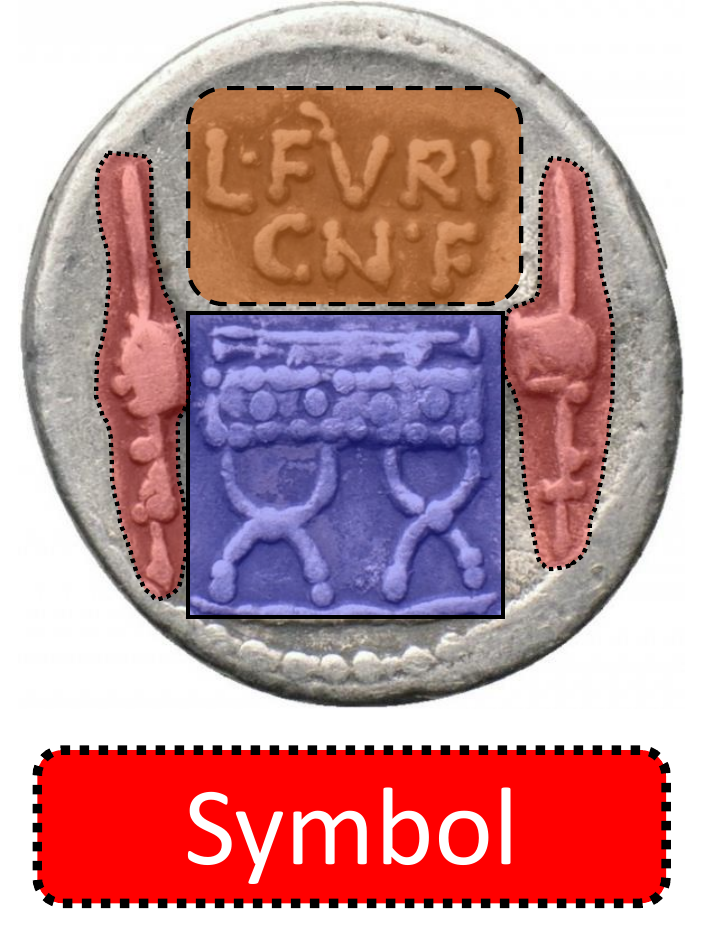}&  
    \includegraphics[width=0.33\linewidth, keepaspectratio]{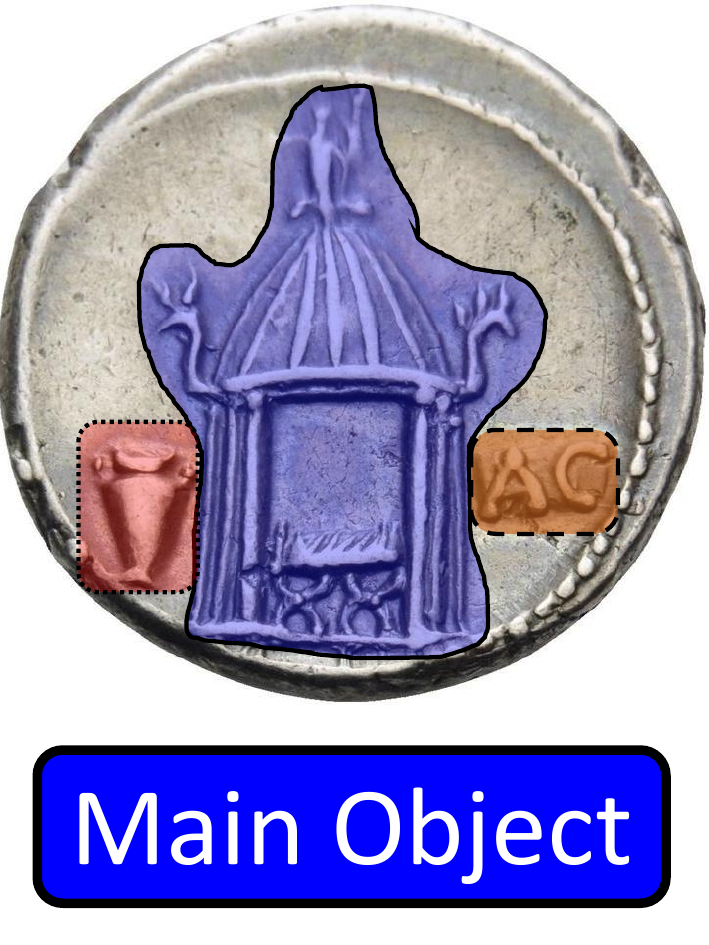}&  
    \includegraphics[width=0.33\linewidth, keepaspectratio]{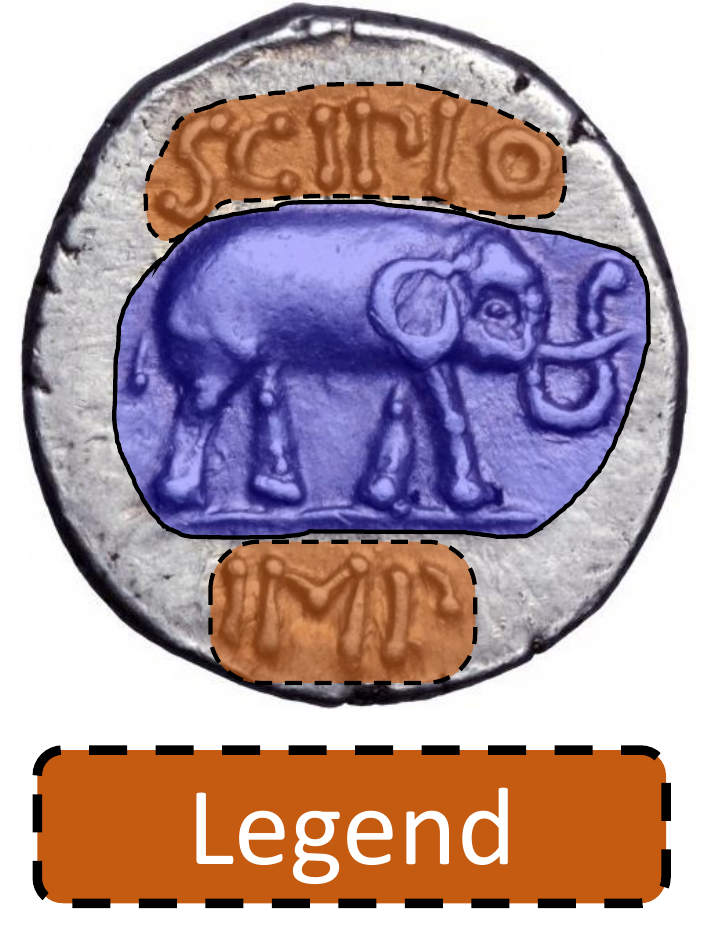}\\
    
 \end{tabular}
 }
\end{center}
\caption{\textbf{Dataset Challenge:} Variations in the anatomy of the reverse motifs due to the positions of the symbol, main object, and legend.}
\label{fig:1a}
\end{figure}

In this paper, we strive to facilitate ancient coin recognition by introducing one of the largest and the most diverse datasets presented so far. The categorization of our Roman Republican Coin Dataset, which we call RRCD, is based on the \textit{main object} shown on its reverse side. The object represented on a coin can have many forms, such as a person, instrument, animal, and building, to give a few examples. The object is the main element for coin classification in addition to the coin legend and smaller auxiliary symbols. We call these visible marks as motifs. Since there is a huge variation among the positions of these motifs on the Roman Republican coins, the task of image-based coin classification is very challenging and non-trivial. Exemplar images are shown in Figure~\ref{fig:1a} where the variations in the anatomy of the coins are evident. Exacerbating such inter-class differences, severe intra-class inconsistencies are commonly found in the ancient coins due to manual minting, abrasions, missing parts, intentional deformations, usage,  rust, and patina.  

Our dataset consists of 228 object classes minted on the reverse side, such as quadriga, griffin, elephant, and many more. We show that a domain adapted neural network model trained with a comparably a small dataset can retrieve the object class with high reliability. We believe that this work is a meaningful step towards a better semantic understanding of ancient coins, as the recognition of their essential elements is key to the ancient coin classification task. Training coin images with a coin ID from a reference classification scheme such as \cite{crawford1974roman} is impractical for large-scale systems due to the vast amount of classes and the burdensome effort to collect training samples, especially for rare coin classes. Therefore, a system that can recognize the essential elements like the main object or legend would provide a semantically meaningful output that can also easily be mapped to possible coin classes in case it links into a respective ontology.

To iterate, the classification problem we tackle has these inherent challenges: 
\begin{itemize} 
\item Huge appearance variations induced by the reverse motif of the ancient coins anatomies, abrasions, and the number of coin classes, 
\item Absence of a large-scale dataset established under strict numismatics guidelines, and 
\item Lack of sufficient exemplary images for a greater proportion of coin classes to train or test the classification model. 
\end{itemize}

Our contributions towards this fine-granular coin recognition task can be summarized as: 
\begin{itemize} 
\item We develop a novel and domain-specialized neural network model called the \enquote{CoinNet}. 
\item We introduce the largest and most diverse dataset of the Roman Republican coins, called RRCD, collected from three specialized numismatics resources under strict and coherent guidelines. Our dataset is available publicly.  
\item We demonstrate that our solution's generalization ability outperforms existing CNN models on completely disjoint test sets that accommodate coin classes having fewer exemplary images. 
\end{itemize}

The remainder of this paper is structured as follows. Section~\ref{sec:related work} provides a literature overview. Section~\ref{sec:new_dataset} introduces our novel large-scale image dataset of the Roman Republican coins. Section~\ref{sec:coins recognition} explains the architecture of the CoinNet. Section~\ref{sec:results} reports the results and ablation study. 

\section{Related Work} 
\label{sec:related work} 

The high applicability of the image-based recognition of coins~\cite{reisert2006efficient} and currency notes~\cite{feng2014automatic} makes it a non-trivial research domain. Consequently, the earlier work on coin classification approaches targeted modern-day coins, which is comparably straightforward since the use of modern technology for coin manufacturing
ensures a uniform visual appearance concerning shape, depictions, and legend
on the obverse and reverse sides. As a result, relatively simple image analysis schemes based on traditional approaches such as geometric shape features~\cite{nolle2003dagobert},
gradient~\cite{reisert2006efficient} and eigenspace~\cite{huber2011identification} achieved notable classification rates on modern coin datasets with as much as 2270 coin classes. Despite their success on modern coins, these approaches were shown to perform poorly on the task of ancient coin classification~\cite{zaharieva2007image}. As a remedy, attempts on ancient coin classification incorporated additional analysis on visual depictions such as portrait recognition~\cite{arandjelovic2010automatic}, object recognition~\cite{anwar2015ancient,anwar2013bag}, and legend recognition~\cite{kavelar2012word}.

Generally, strategies for ancient coin classification constitute two main groups. The first one uses local feature matching techniques. Local features capture image variations in a local neighborhood, and the set of such features calculated over an entire image provide a discriminating representation of that particular image. Local features allow calculation of the similarity of two images by measuring, for instance, Euclidean distance between the corresponding local features. The second group of methods for ancient coin classification uses supervised learning algorithms. The parameters of these algorithms are derived in an offline training process with the help of training image datasets. The learned model is then used to predict the class for a test image.

The success of supervised learning mainly comes from the availability of abundant data for the offline training phase. However, in the case of ancient coins, the prevalent problem is the absence of training data due to their rareness and diversity, thereby leading to low recognition rates~\cite{zambanini2014classifying}. In comparison, the feature matching based techniques neither involve an offline training process nor require a large number of exemplary images. Even with three or four samples per class, the feature matching substantially outperforms the supervised learning methods~\cite{zambanini2014classifying}. Nevertheless, the online feature matching, as well as the search process it involves, make the feature matching methods computationally intensive. Besides, the complexity increases proportionally with the number of classes in the dataset~\cite{zambanini2014classifying}.

The first exclusive method for ancient coins~\cite{kampel2008recognizing} uses a combination of local feature descriptors~\cite{lowe1999object,bay2008speeded,mikolajczyk2005performance} to perform an exemplar-based classification. Zambanini and Kampel~\cite{zambanini2012coarse} apply dense correspondence-based feature matching called SIFT flow~\cite{liu2011sift}. To improve the quality of local features matching, Zambanini \etal~\cite{zambanini2014classifying} employ the geometric consistency of the matched features. Similarly, a more customized descriptor for ancient coin classification called Local Image Descriptor Robust to Illumination Changes (LIDRIC)~\cite{zambanini2013local} is proposed to alleviate the sensitivity to illumination changes. To sum up, in the absence of training data, the feature matching-based methods achieve acceptable classification rates. However, they are not easily scalable to more extensive datasets. They disregard the inherent domain-specific knowledge, which is extremely important from a numismatics perspective to make the classification task complaint with the standard reference books in this subject.

Leveraging upon the numismatic knowledge, the second group of methods uses machine learning algorithms for ancient coin classification. For instance, recognition of legends on the obverse sides of Roman Imperial coins is used for their classification~\cite{arandjelovic2012reading}. The legend is assumed to be located along the coin border and is curvature normalized by a log-polar transformation. Due to this assumption, their method performs poorly on Roman Republican coins where the legend's location is not fixed~\cite{KavelarZK14}. Consequently, Kavelar~\etal~\cite{kavelar2012word} performs legend recognition of Roman Republican coins using SIFT features with a Support Vector Machine (SVM). The legends carry rich information in terms of alphabets and numbers, thus making them an excellent cue for coin classification. However, they suffer more wear and tear on the coins due to their detailed nature, which makes them less attractive and impractical for coin classification~\cite{schlag2017ancient}. Another visual cue used for ancient coins classification is the obverse side portrait~\cite{arandjelovic2010automatic,kim2014improving}. However, the semi-frontal portraits on the obverse side have less inter-class variations~\cite{schlag2017ancient}. Also, like legends, the portraits are more likely to lose their details with erosion. Anwar~\etal~\cite{anwar2013bag,anwar2015ancient} utilizes reverse motifs recognition for ancient coin classification where the spatially enriched Bag-of-Visual-Words (BoVWs)~\cite{csurka2004visual} model represents the images. 

Unlike legends and portraits, the reverse motif is a discriminative visual cue that is less affected by wear and tear. Besides, a given reverse motif can be shared by coins of multiple classes. Therefore, the search space for the class of a given query coin image is aptly reduced by recognizing its reverse motif. This aspect makes the reverse motif-based coin classification coarse-grained that can further be refined by fine-grained classification methods~\cite{zambanini2014classifying}.

A comprehensive review of recent deep learning methods is out of the scope of this paper. Still, we like to mention that the convolutional neural networks have already been applied in the field of digital humanities. Kim and Pavlovic~\cite{kim2016discovering} have used CNNs to recognize the prominent visual cues on the ancient coins and later utilized them for classification. Similarly, Schlag and Arandjelovi{\'c}~\cite{schlag2017ancient} have used CNNs for portrait recognition on the obverse side to classify the ancient Roman Imperial coins. However, the process of massive data collection, such as the sources, methods, and guiding principles, is of extreme importance, especially when it comes to ancient objects such as coins. Such a procedure is not outlined in existing CNN based methods, which makes them unreliable.

On the other hand, we explicitly elaborate on the process of collection of the largest dataset of the Roman Republican coins. Our sources of coin images are among the most reliable ones in this field. Lastly, our guiding source in the data collection is the standard reference book by Crawford~\cite{crawford1974roman}, which is considered as the utmost authority on the Roman Republican coins by the numismatists. A detailed description of the data collection is explained in the next section.  

\section{RRCD - Roman Republican Coin Dataset} 
\label{sec:new_dataset} 
The primary motivation of research on ancient coin recognition is to support the manual coin classification efforts by reducing the labor time involved in the process. To make the best use of the invaluable domain expertise, the recognition task should be steered by the standard reference books of numismatics. However, this critical aspect is often overlooked in most published work on coin classification except for a few~\cite{anwar2015ancient}. Similarly, the use of smaller image datasets to evaluate the proposed coin classification methodologies leads to unrealistic and ungeneralizable approaches. Even in the recently introduced and relatively larger datasets~\cite{schlag2017ancient}, the coin images are categorized into different grades without involving domain-specific knowledge, thus creating ambiguity about the feasibility of the solutions. 

Our focus is on the gold and silver coinage of the Roman Republican era (BC 280/225-27) since there is a comprehensive standard reference work by Crawford~\cite{crawford1974roman}, which is still accurate today, with only minor modifications \cite{hollstein1993stadtromische,woytek}. Crawford's work assigns 550 distinct reference numbers, many comprising different denominations or typological variations. By consolidating all the variants, the actual number of possible combinations might exceed 2000.

\begin{figure}[!t]
\centering
\includegraphics[width=0.9\textwidth]{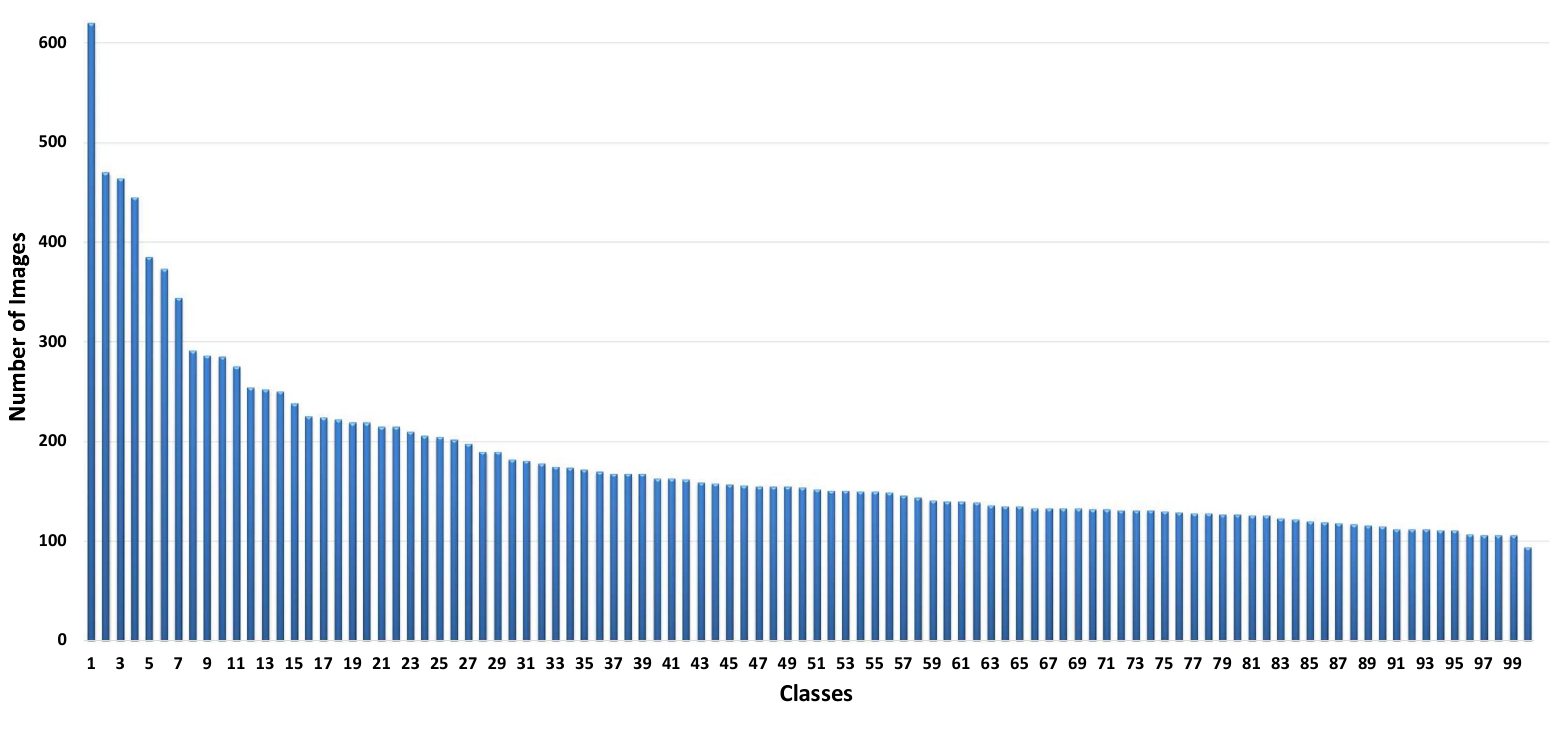}
\caption{\textbf{Number of Classes:} Per-class image counts in the dataset.}
\label{fig:1}
\end{figure}

Based on Crawford's work, we collect the most diverse and extensive image dataset of the reverse sides. For most of the Roman Republic coin classes, the obverse side depicts more discriminative information than the observe side~\cite{crawford1974roman}. Our dataset has 228 motif classes, including 100 classes that are the main classes for training and testing, which we call the \textit{main dataset} RRCD-Main. The images of the additional 128 classes constitute the \textit{disjoint test set}, RRCD-Disjoint, which we allocate to assess the generalization ability of our models. Therefore, the training and testing can be evaluated on completely disjoint datasets. The number of images per class in the RRCD-Main is shown in Figure~\ref{fig:1}, while a comparison with the existing available reverse side datasets in the literature is given in Table~\ref{tab:datasets}. To the best of our knowledge, RRCD is the most diverse dataset proposed while it is the largest dataset of the Roman Republican coins. 

\begin{table}[t!bp]
\centering
\caption{\textbf{Datasets comparison:} Image datasets of the ancient Roman coins. Imperial is represented by \textbf{RI} while Republican is given by (\textbf{RR}). The dataset classification is performed based on their different visual cues; Obverse side (\textbf{O}), Reverse side (\textbf{R}), or Legends (\textbf{L}).}
\label{tab:datasets}
\begin{tabular}{|l|c|c|c|c|c|}
\hline
\multicolumn{1}{|l|}{\textbf{Datasets}} & \multicolumn{1}{l|}{Images} & \multicolumn{1}{l|}{Image Size} & \multicolumn{1}{l|}{Visual Cues} & \multicolumn{1}{l|}{Classes} & \multicolumn{1}{l|}{Era} \\ \hline
\cite{schlag2017ancient}        & 49,571 &        -       & O      & 83 & RI \\ \hline
\cite{kim2016discovering}       & 4,500  & 350$\times$350 & O,R    & 96 & RI \\ \hline
\cite{anwar2015ancient}         & 2,224  & 480$\times$480 & R      & 29 & RR \\ \hline
\cite{kim2014improving}         & 2,815  & 256$\times$256 & O      & 15 & RI \\ \hline
\cite{zambanini2014classifying} & 600    & 150$\times$150 & R      & 60 & RR \\ \hline
\cite{zambanini2013improving}   & 464    & 384$\times$384 & O,R,L  & 60 & RR \\ \hline
\cite{zambanini2012coarse}      & 180    & 150$\times$150 & R      & 60 & RR \\ \hline
 \cite{kavelar2012word}         & 180    & 384$\times$384 & L      & 35 & RR \\ \hline
\cite{arandjelovic2010automatic}& 2,326  & 250$\times$250 & O      & 65 & RI \\ \hline
\cite{kampel2008recognizing}    & 350    &  -             & O,R    & 3  & RR \\ \hline
\textbf{Our: RRCD}               & \textbf{18,285} & \textbf{448}$\times$\textbf{448} & \textbf{R} & \textbf{228} & RR \\ \hline
\end{tabular}
\vspace{5mm}
\end{table}

Nonetheless, during image search, we use the reference number given to each coin class by Crawford. The retrieved coin images and the textual descriptions of their obverse and reverse sides are then cross-matched with the standard ones. This allows for a coherent and unambiguous collection process of image data-driven by domain-specific knowledge. We do not perform an explicit categorization of the collected coin images based on their grades. 
However, the deteriorated coin images, where the reverse motif is challenging to be distinguished by the domain experts, are discarded.

\subsection{Composition of RRCD-Main} 

The images in the main dataset RRCD-Main are collected from three different reliable sources. The images from the Vienna Museum of Fine Arts and the British Museum London are captured in a controlled environment, due to which both the image resolution and imaging conditions are of high quality. Furthermore, the coin specimens are in fair condition and do not exhibit extreme visual deterioration. Similarly, the third source is an online ancient coin auction website where both the quality and the coin specimen's condition vary. Nonetheless, images from all the sources face extra variations due to irregular illumination caused by the non-rigid nature of the coins. Following is a brief description of the image data from each source. 

\noindent 

\textbf{The Vienna Museum of Fine Arts:} The stock of material from the Roman Republic in the Coin-Cabinet in Vienna is among the biggest in the world. It comprises about 3900 coins. The ILAC project~\cite{kavelar2013ilac} collected the image dataset of these coins with a uniform background. However, orientation differences exist between the coin images as they are not photographed under their canonical orientations based on their central reverse motifs. We acquired 1416 images from this source. 
\noindent 

\textbf{The British Museum:} An extensive collection of ancient coins is owned by the department of medals and coins at the British Museum. In our dataset, we use 2376 images of the Roman Republican Coins of the British Museum.  
\noindent 

\begin{figure}[t]
\begin{center}
\includegraphics[width=\textwidth]{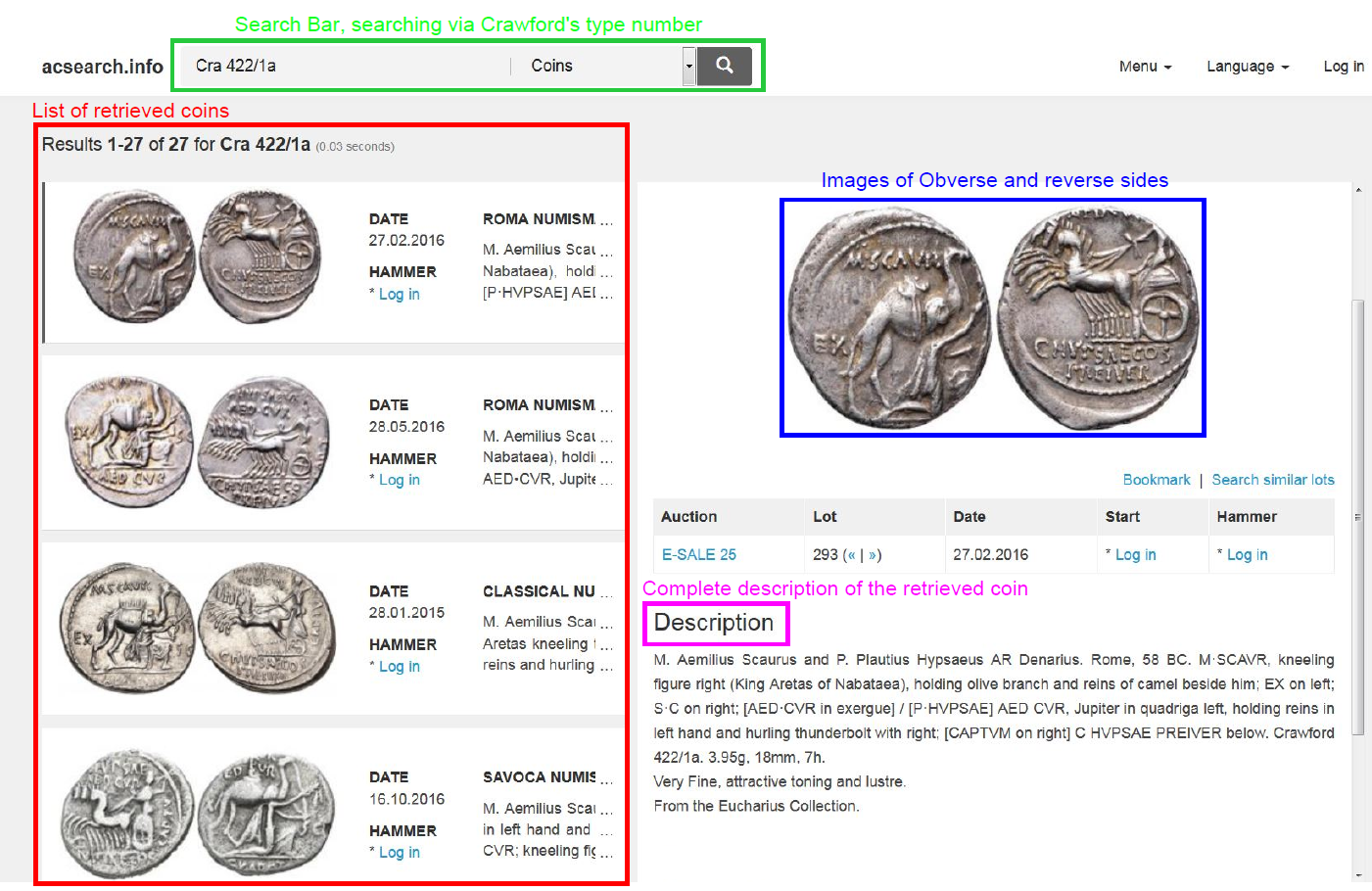}
\end{center}
\caption {\textbf{Search Process:} A snapshot of the acsearch image search process}
\label{fig:acsearch}
\end{figure}

\textbf{The ACSearch:} This is an online auction website of ancient coins including those of the Roman Republican era. For any coin at the auction, the images of both reverse and observe sides with their respective descriptions are provided. The information contains the type of the coin given by Crawford, the issuer, the date of issuance, and explanations of the objects, scenes, or persons depicted on each side. A snapshot of the website is shown in Figure~\ref{fig:acsearch}, where various parts of the page showing different information are highlighted. 

\begin{figure}[t!bp]
\begin{center}
\includegraphics[width=\textwidth]{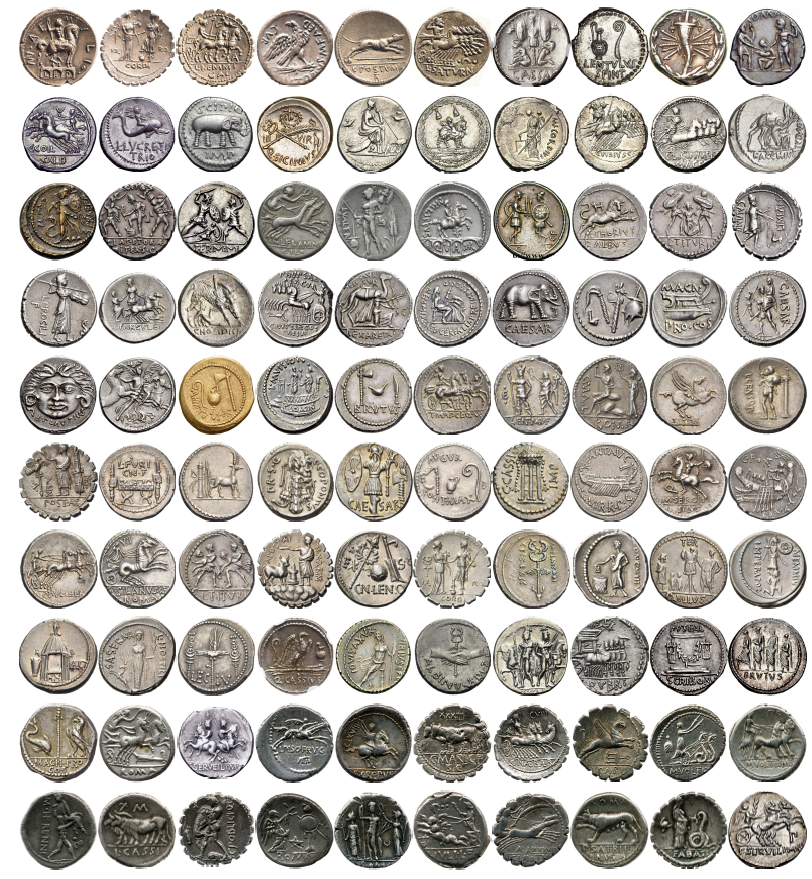}
\end{center}
\caption{\textbf{Representative images:} Samples images of the 100 classes that constitute the RRCD-Main. 
}
\label{fig:2}
\end{figure}

A coin can be searched via the website's search bar using the keywords from the description, such as the type number or the object displayed on the reverse side. For a search coherent with the standard reference book, we used the type numbers given by Crawford~\eg~\enquote{Cra. 422/1a}. This results in a list of coin image retrievals along with descriptions. For a uniform collection, we match the images with their descriptions and download only those images that are in complete agreement with their records. We also cross-check the retrieved information with the descriptions given in the reference book. Exemplar reverse side images of the RRCD-Main classes are shown in Figure~\ref{fig:2}.

\subsection{Composition of RRCD-Disjoint} 

In many cases, the main object on the reverse sides of the Roman Republican coins is shared by multiple coin classes. However, the object style and the additional information on the reverse motifs such as the symbols and legends make the coin classes different from one another. To include images of all those styles in the RRCD-Main is impractical, mainly due to the lack of their images or the rarity of the specimen themselves~\cite{zambanini2012coarse}. Due to such constraints, an image-based coin classification solution should be robust to variations in object styles. If trained on one set of object styles, the framework should be generalizable enough to recognize other styles. 

To investigate the performance of our proposed CoinNet and assess its generalization, we select the predominant objects found on the reverse motifs of the Roman Republican coins; namely, \enquote{biga} (two-horse chariot), \enquote{quadriga} (four-horse carriage), and \enquote{curule chair}. Out of 100 coin classes of RRCD-Main that we collected from three different sources, 12 coin classes have biga as the main object, four coin classes show quadriga, and there is only one coin class where the curule chair is minted. The depiction of main objects varies from each other depending on their styles, additional symbols, and legends.   
  
\begin{figure}[t!bp]
\begin{center}
\includegraphics[width=\linewidth]{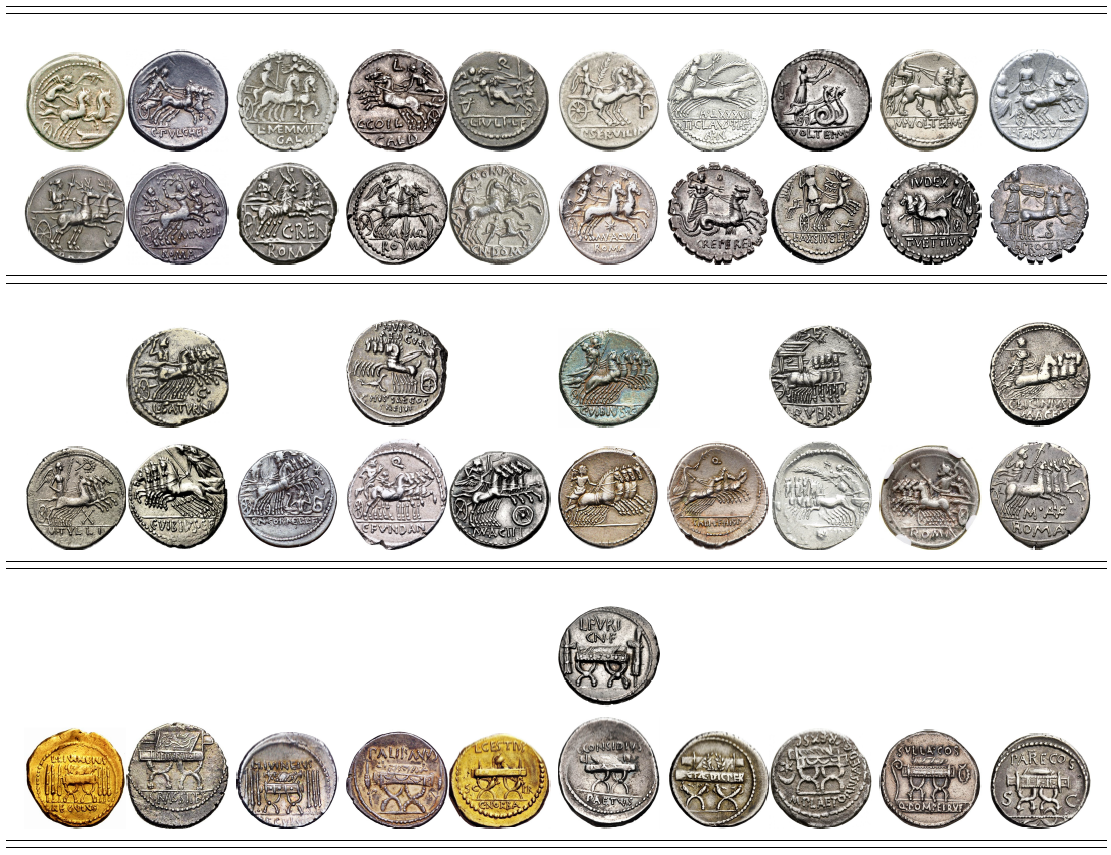}  
\end{center}
\caption{\textbf{Disjoint image set:} The first row in each partition (separated by double lines) shows images of the coin classes included in the RRCD-Main while the second row shows exemplar images of some of the coin classes included in the RRCD-Disjoint test set. Each partition depicts a separate main object; biga, quadriga, and curule chair, respectively. 
Since the same main object is minted in different styles with different additional symbols and legends, we treat each column as a separate class. 
}
\label{fig:BIGA_examples}
\end{figure}

We collect another 700 images of 81 coin classes where biga is minted in styles different from those of the RRCD-Main. Similar sets of 111 images for 12 curule chair classes and 344 images of 35 quadriga classes are collected too (total 128 classes). We call the combination of all these image datasets as the \enquote{the disjoint test set} RRCD-Disjoint because they are only used to test the coin classification framework that is trained on the main dataset RRCD-Main. The exemplar images of the coin classes from RRCD-Main, along with the representative images of some of the classes of RRCD-Disjoint, are shown in Figure~\ref{fig:BIGA_examples}. The differences in styles between the coin classes in RRCD-Main and those in RRCD-Disjoint can clearly be noted. For instance, in the case of biga and quadriga, the following are the main differences: 

\begin{enumerate} 
\item Bigas have different animals such as horses, stags, lions, snakes, goats, seahorses, and Centaurus.
\item The animals also vary due to their moving styles \ie, they are either walking, running or galloping 
\item The chariots are either moving towards the right or left 
\item The persons driving the chariots are depicted differently. For instance, they vary from one another due to the objects in their hands. 
\item There are additional symbols associated with the chariots. 
\end{enumerate} 
Similar differences exist for quadriga and curule chair where it is either minted in a different style or have different symbols and legends.  

\section{CoinNet: Proposed Coin Recognition Network}
\label{sec:coins recognition}

In coin recognition task, we aim to predict the most likely outcome $\hat{c}$ for any given image $I$, which can be expressed as:
\begin{equation}
\hat{c} = \argmax_{c \in C} p( c | I, \theta),
\label{eq:arg}
\end{equation}
where $\theta$ are the network parameters and $C$ is the set of classes. It needs to be regarded here that Eq.~\ref{eq:arg} takes image $I$ to predict the class label, while we extract image embeddings (feature maps) $\alpha_I$ and $\beta_I$ from the input image using off-the-shelf convolutional neural networks. Therefore, Eq.~\ref{eq:arg} can be rewritten as 
\begin{equation} 
\hat{c} = \argmax_{c \in C} p(c|\alpha_I, \beta_I, \theta). 
\label{eq:arg_feat} 
\end{equation} 
Our purpose is to exploit a joint representation by employing an appropriate pooling operator $\phi(\cdot)$ which can encode the relationship of Eq.~\ref{eq:arg_feat} between feature maps $\alpha_I$ and $\beta_I$. 

\subsection{Compact Bilinear Pooling} 

The bilinear models are introduced by \cite{tenenbaum2000separating} and received remarkable performance improvement in computer vision and image processing tasks. However,  bilinear presentations are impractical due to many reasons: 1) the number of parameters becomes very high, 2): the features stored in memory for retrieval or deployment requires TeraBytes of storage, 3):  the processing for matching and domain adaptation requires feature concatenation, which stresses memory and storage and 4): scenarios such as few-shot~\cite{Sun_2019_CVPR} and zero-shot learning~\cite{Xie_2019_CVPR} becomes challenging.  Here, we first provide the formulation of the bilinear representation and then introduce its compact form. 

In our case, the bilinear model $M$ is obtained by taking the outer product of the two vectors ($\alpha_I~\in~\mathcal{R}^{m_1}$ and $\beta_I~\in~\mathcal{R}^{m_2}$) as  
\begin{equation}
z_I = M\big( vec(\alpha_I~\otimes~\beta_I) \big),
\label{Eq:linear_model}
\end{equation}
where $vec(\cdot)$ converts the matrix into a vector form, \ie vectorize the product. The bilinear model is effective as it computes each interaction between the encoded vectors; however, it is computationally expensive, as mentioned earlier. Let us consider an example where $m_1$=2048 and $m_2$=2048 with $C$=100 output classes (\ie $z_I \in \mathcal{R}^{100}$) will result in a high dimensional representation with the model composed of one billion parameters.

To avoid the outer product in bilinear models and project the representation onto a lower-dimensional space, we employ the Compact Bilinear Pooling (CBP) of~\cite{gao2016compact} and~\cite{fukui2016multimodal}, where both propose to utilize the Count Sketch Projection function~\cite{charikar2002finding} which projects a vector $x \in \mathcal{R}^{n_1}$ to $y\in \mathcal{R}^{n_2}$. Count Sketch Projection randomly draws two vectors $u \in \{-1,1\}^{n_1}$ and $v \in \{1,\ldots,n_2\}^{n_1}$ from a uniform distribution, while these drawn vectors remain constant for the future invocations. The mapping function $v$ maps the $i^{th}$ index of $x$ to the $j^{th}$ index of $y$, initialized as zero. For every element $x[i]$ its destination index $j = v[i]$ is looked up using $v$; and then $x[i]$ is added to $y[j]$. This technique helps to reduce the number of parameters in the model due to the projection of the outer product (bilinear representation) to a low-dimensional space.

According to~\cite{pham2013fast}, the computation of outer product can be circumvented by taking the convolution of the count sketches as  
\begin{equation}
\phi(\alpha, \beta, u, v) = \phi(\alpha, u, v) \ast \phi(\beta, u, v),
\label{eq:count_sketch_convo}
\end{equation}
where $\ast$ is the convolution operator. Furthermore, according to the convolution theorem, the element-wise multiplication in the one (frequency) domain is equal to convolution in the other (spatial) domain. Therefore, Eq.~\ref{eq:count_sketch_convo} can be rewritten as 
\begin{equation}
\phi(\alpha, \beta, u, v) = \mathrm{F}^{-1} (\mathrm{F}(\phi(\alpha, u, v)) \odot\mathrm{F}(\phi(\beta, u, v))),
\label{eq:count_sketch_fft}
\end{equation}
where $\odot$ is the element-wise multiplication operator, and $F$ is the Fourier transform function. In the next section, we describe the convolutional neural network segment of our algorithm.

\subsection{Proposed Architecture}
Our model encodes the input image to extract feature maps and then merges them via the compact bilinear pooling algorithm. The problem is treated as a multi-class classification task with 100 possible outcomes. As a first step, the images are resized to 448$\times$448 and encoded using the two popular CNN networks \ie DenseNet161~\cite{huang2017densely} and ResNet50~\cite{he2016deep} having 161 and 50 convolutional layers trained on ImageNet dataset~\cite{deng2009imagenet}. The features are collected from the network before the final fully-connected layer without applying the average pooling resulting in a 14$\times$14$\times$2048 feature map. Suppose DenseNet161~\cite{huang2017densely} and ResNet50~\cite{he2016deep} are denoted by $\Upxi$ and $ \Omega$, 
  
\begin{align*}
  \alpha_I &= \Upxi (I),\\
  \beta_I &= \Omega (I).
\label{eq:features}
\end{align*}
We fuse these feature maps $\alpha_I$ and $\beta_I$ using CBP to obtain a better representation. Then we apply a group of residual blocks to the fused features to learn the joint representation. We also perform $\ell_2$ normalization on the 2048-D vector obtained from the residual group. 

\begin{figure}
\begin{center}
\includegraphics[width=1\textwidth]{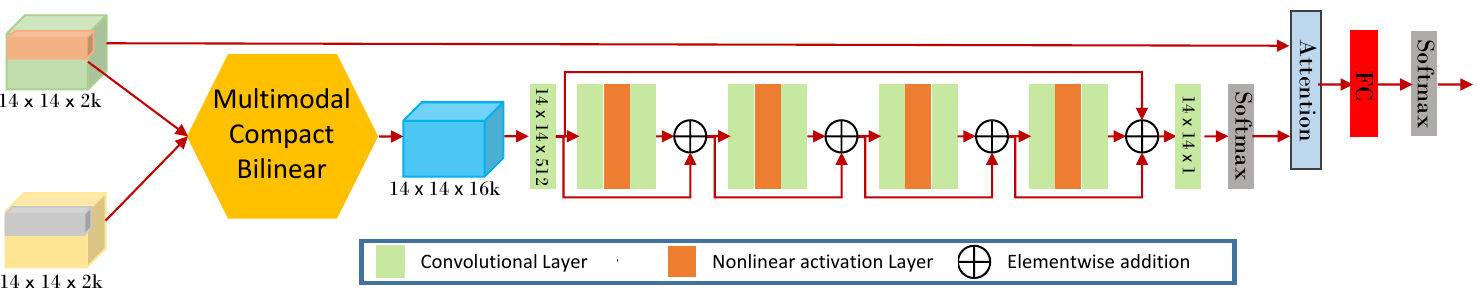}
\end{center}
\caption{\textbf{CoinNet:} Our model highlighting the Compact Bilinear Pooling, residual blocks, skip connections, and feature attention. The green and yellow cubes indicate the embedded features via CNN networks.}
\label{fig:net}
\end{figure}

\vspace{2mm}
\noindent
\textbf{Attention:} Recently, attention has been investigated in many computer vision applications, \eg image captioning~\cite{xu2015show}, super-resolution~\cite{zhang2018image} and visual question answering~\cite{yang2016stacked}. In our model, we also incorporate soft-attention to integrate spatial information. As presented in Fig~\ref{fig:net}, we employ one convolutional layer to extract features to emphasize the salient features. Moreover, we apply softmax to predict each grid location's attention weights to generate normalized soft attention maps. To get the visual representation, the attention map is summed with the spatial feature vectors $\alpha_I$ and $\beta_I$. As a final step, a fully-connected layer is employed to obtain the number of outputs equal to the number of coin categories.

\vspace{2mm}
\noindent
\textbf{Network loss:}
The output features of the fully-connected layers are passed via the softmax function to normalize the feature values. Moreover, the loss function, we compute the difference between the predicted probabilities and the actual distribution of the class through cross-entropy as

\begin{equation} 
l(p,q) = - \sum_{i=1}^n q_i(y)\log(p_i(y)). 
\end{equation}   

Here, $q_i(y)$ and $p_i(y)$ stands for the true and the estimated probabilities, respectively.  Furthermore, the loss only captures the error on the target class where its value is non-zero because $q_i(x)$ uses one-hot encoded vectors. 


\begin{table}
\caption{\textbf{Quantitative comparison:} Comparison of our method with state-of-the-art methods on train-test split of 30\%-70\%. All results reported as top-1 mean accuracy on the test set.}
\centering
\begin{tabular}{|l|c|c|c|c|c|}
\hline
Algorithms  &BoVWs      & RT        & VGG       &NASNet         & Ours  \\ \hline \hline
Acc.        &70.81\%    & 84.4\%    & 97.4\%    &  97.8\%       & \textbf{98.5}\%\\ 
Precision        & -         &  -        & 0.871     & 0.883         & \textbf{0.907}\%\\ 
Recall        & -         &  -        & 0.903     & 0.914         & \textbf{0.951}\%\\ \hline \hline
\end{tabular}
\label{tab:results}
\end{table}

\section{Experiments}
\label{sec:results}
This section of the paper presents a quantitative and qualitative performance evaluation of our method against state-of-the-art traditional and convolutional neural network algorithms. Firstly, we show the focus of our network on different objects for classification. Then, we investigate the influence of various feature inputs on performance accuracy. Subsequently, we report on the generalization capability of our method. 

\subsection{Experimental Setup} 

In this section, we provide implementation details of our model. We set the filter size of all the convolutional layers as 3$\times$3. We use four residual blocks as a single residual group. The initial learning rate is fixed at 10$^{-2}$, which is reduced after 50 epochs by a factor of 10$^{-1}$. To train the model, we use SGD~\cite{bottou2010large} with a weight decay of 10$^{-4}$. We use 30\% of the data for training, utilizing data augmentation, which includes random rotations and flipping. The model is implemented using PyTorch on a Tesla P100 GPU with 16GB memory.

\begin{figure}[t]
\begin{center}
\resizebox{.8\textwidth}{!}{
\begin{tabular}{c@{ }c@{ }c}
    \includegraphics[width=0.33\linewidth, keepaspectratio]{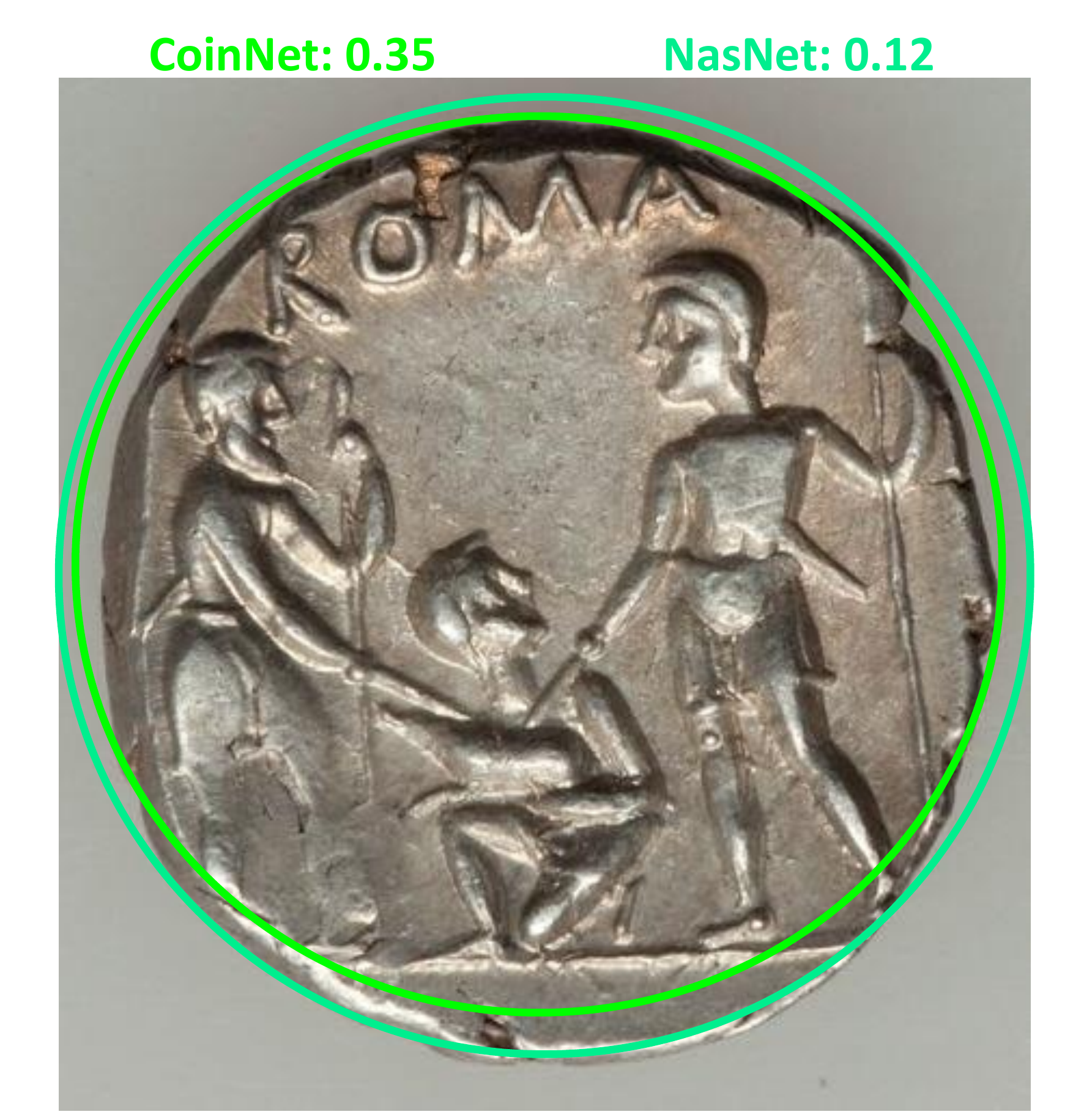}&  
    \includegraphics[width=0.33\linewidth, keepaspectratio]{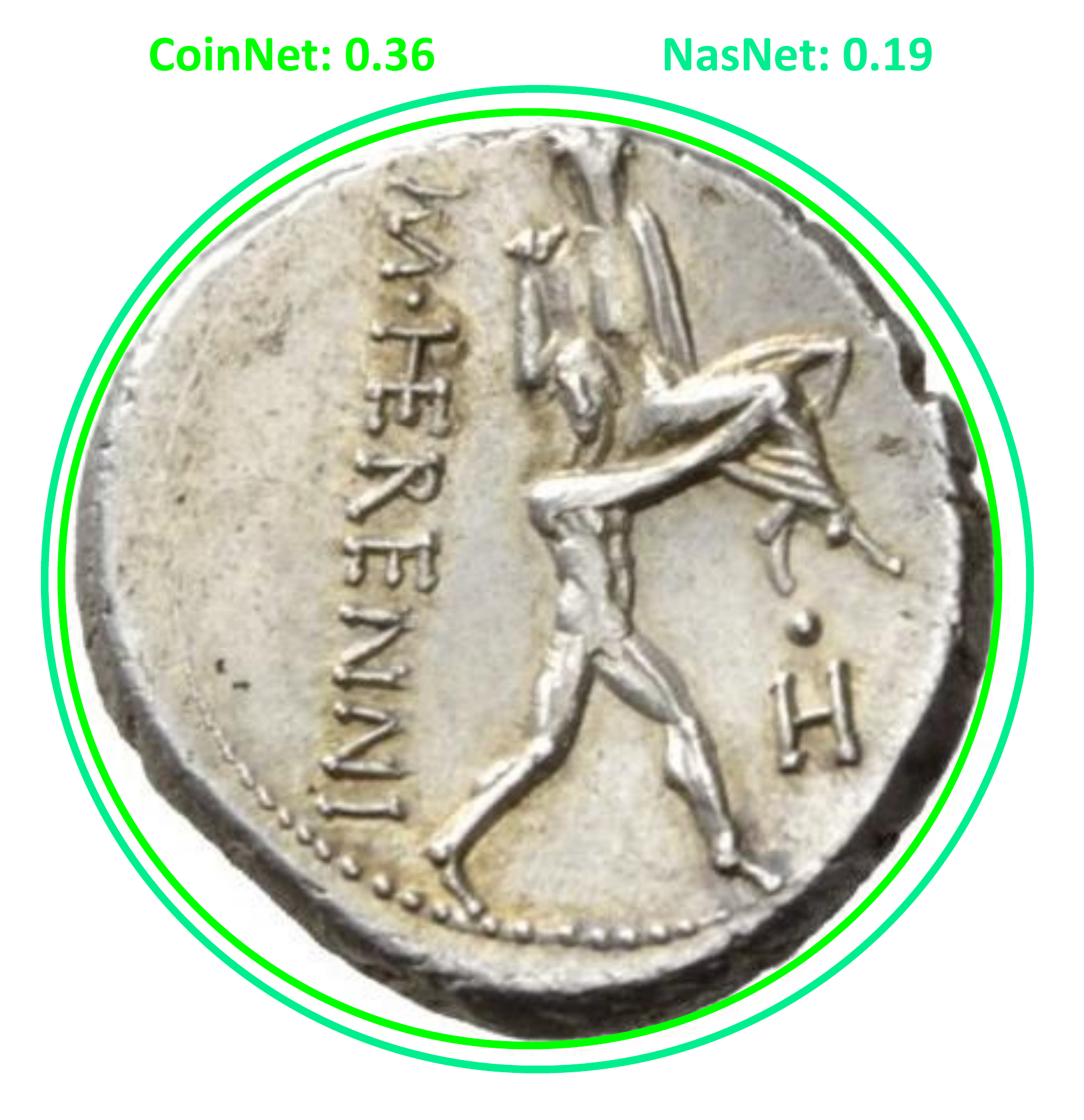}&  
    \includegraphics[width=0.33\linewidth, keepaspectratio]{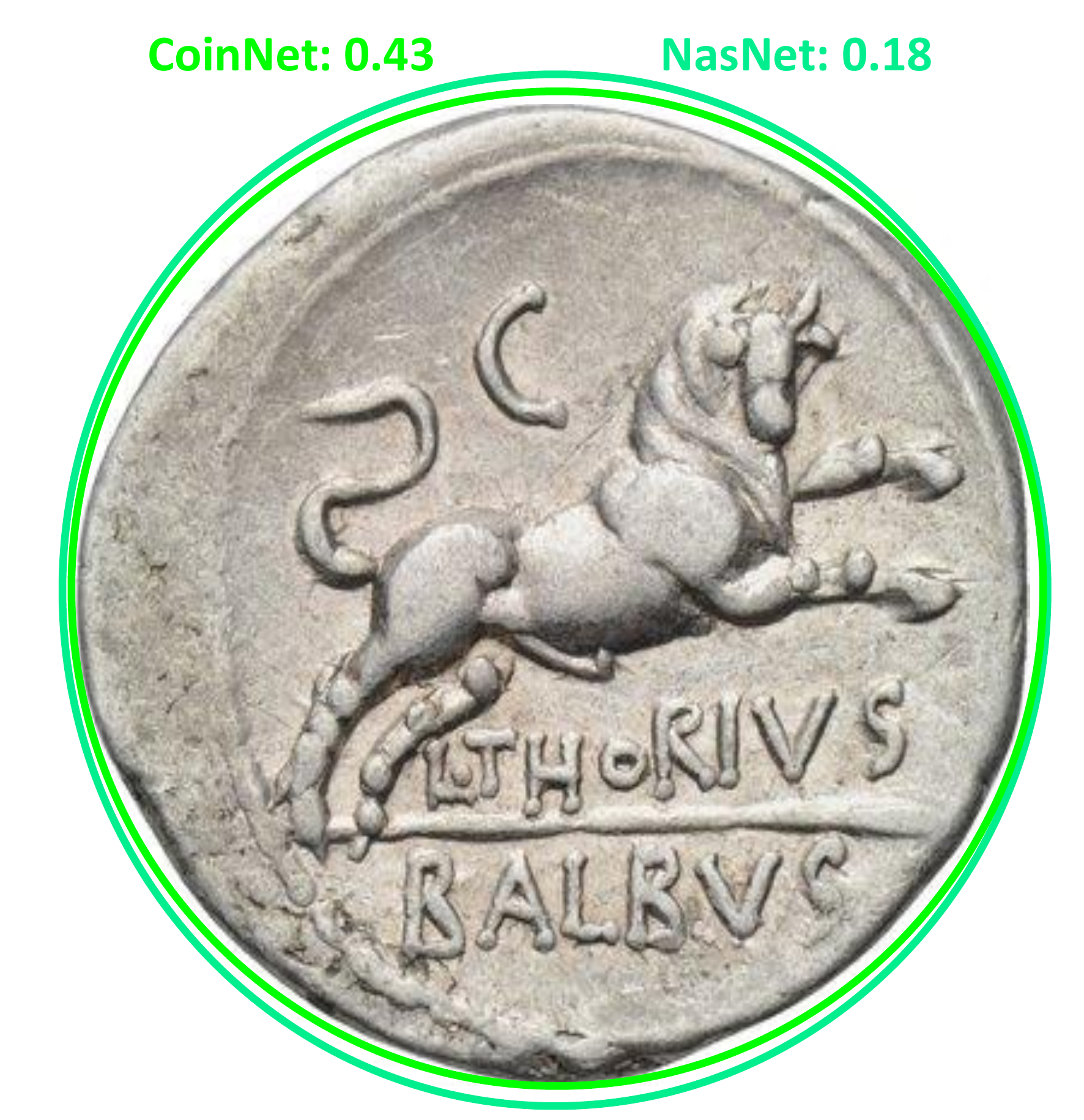}\\  
    Youth and soldiers & Father and son & Charging bull\\
    
    \includegraphics[width=0.33\linewidth, keepaspectratio]{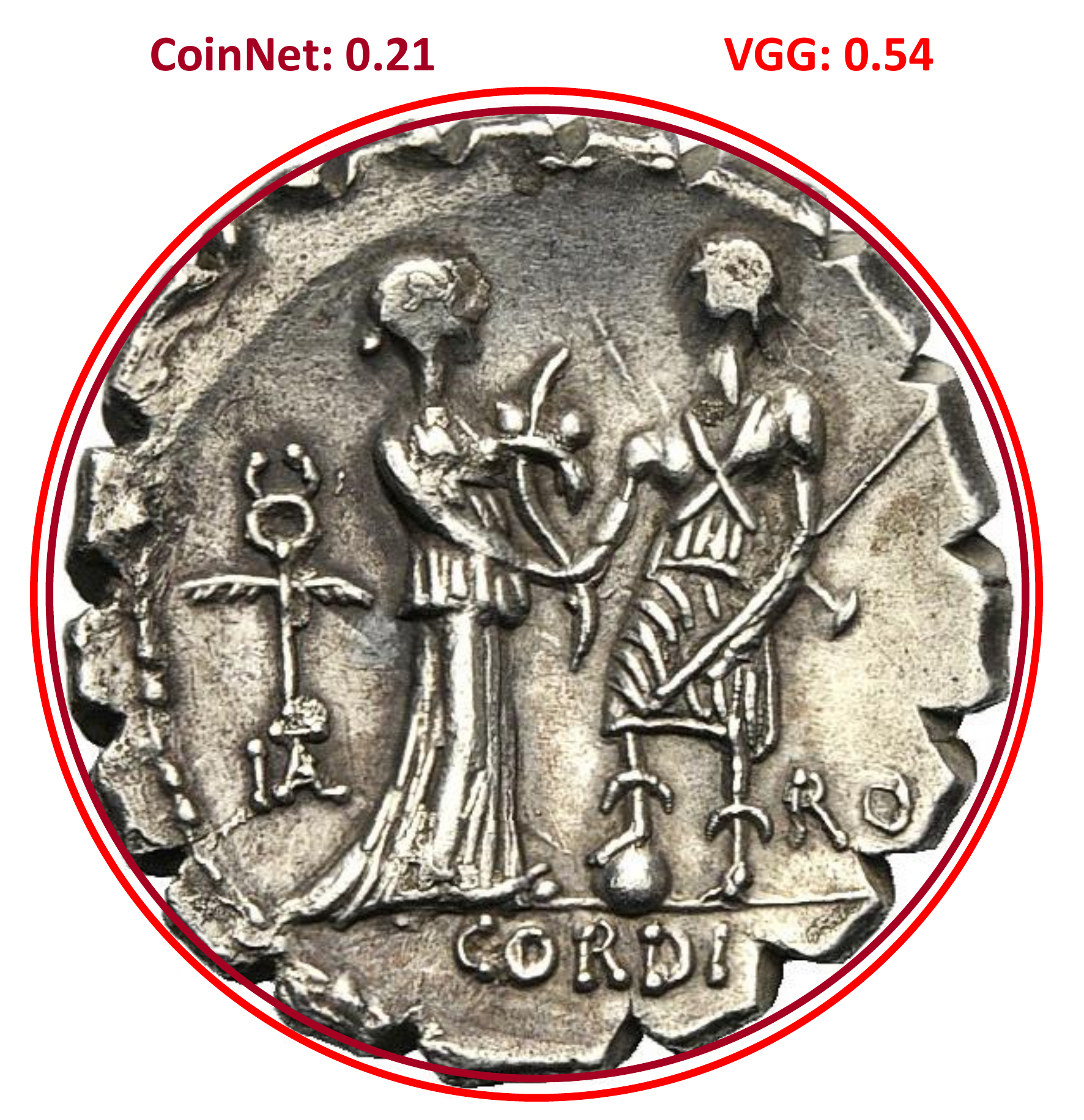}&  
    \includegraphics[width=0.33\linewidth, keepaspectratio]{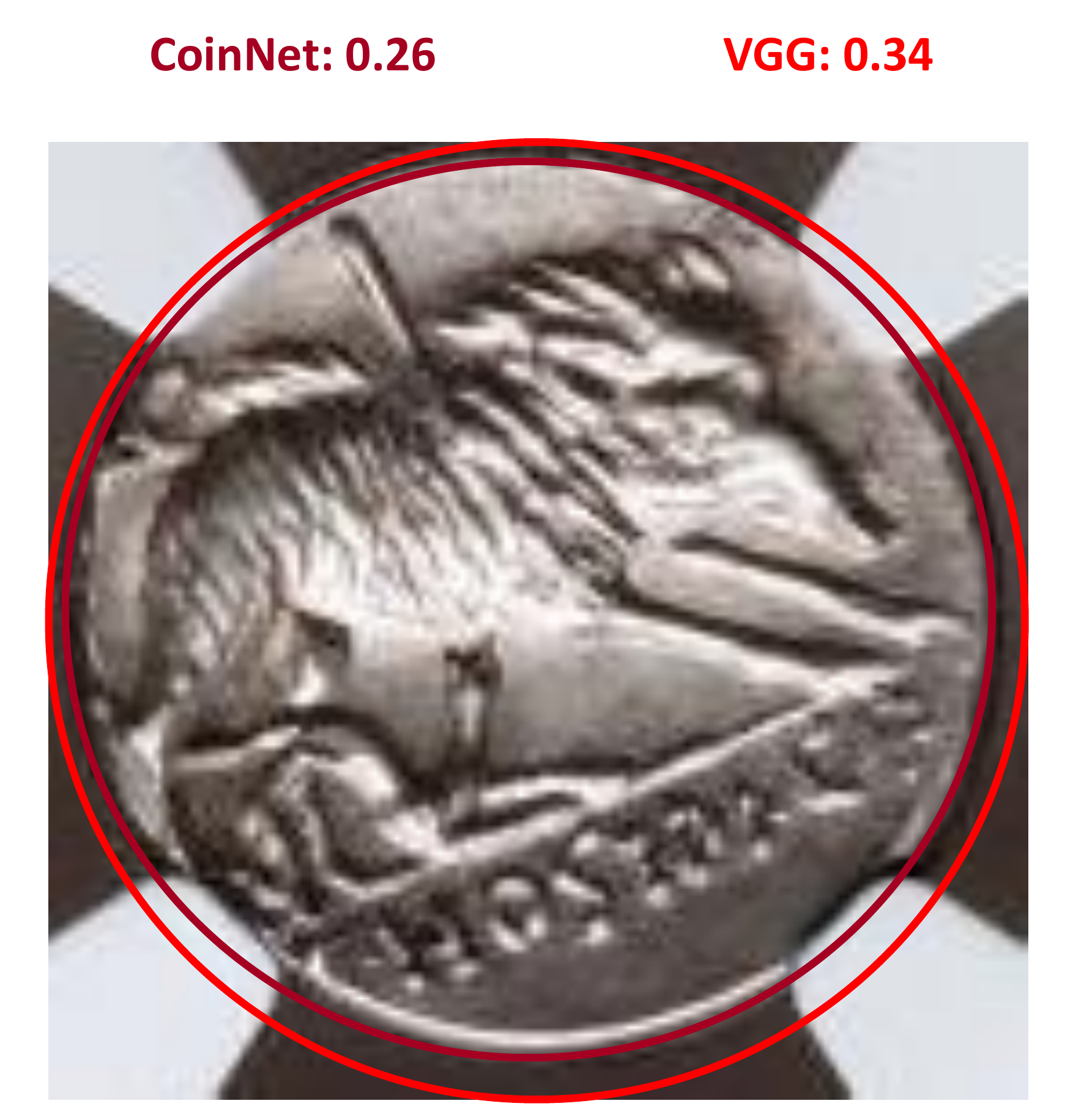}&  
    \includegraphics[width=0.33\linewidth, keepaspectratio]{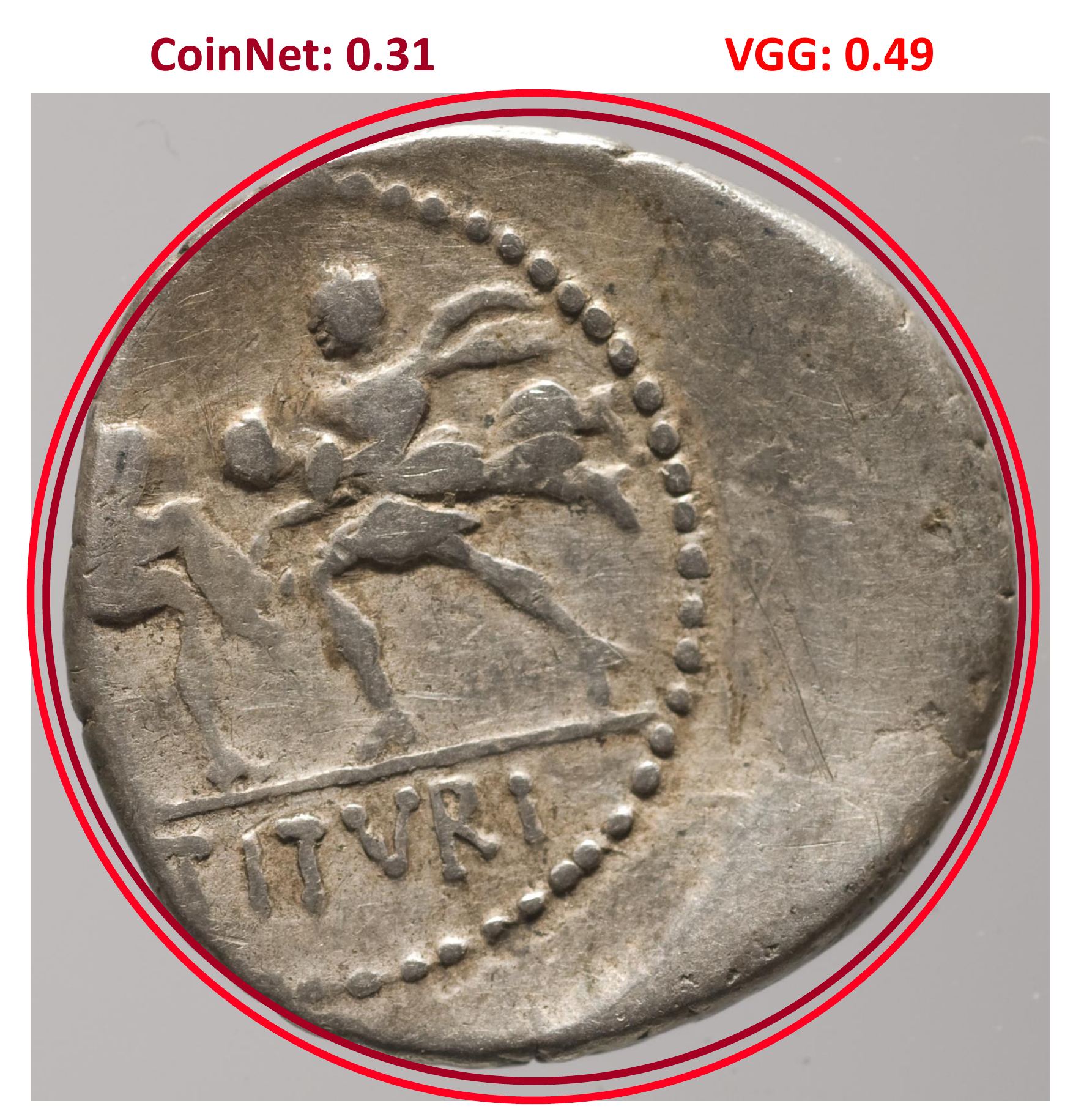}\\  
    Italia and Roma & Wild boar and dog & Soldiers and women\\
 \end{tabular}
 }
\end{center}
\caption{\textbf{Visual comparison:} The correctly classified images are represented with green circles while the wrongly classified ones are in red circles. In the first row, the confidence of the NasNet~\cite{zoph2018learning} is always low, although the model can classify correctly. The second shows the confidence of the VGG~\cite{simonyan2014vgg}, which is consistently high even for wrongly classified classes. }
\label{fig:im_classify}
\end{figure}

\subsection{Quantitative Evaluation}
Until now, Anwar~\etal~\cite{anwar2015ancient} used the largest and most diverse dataset of the reverse side images of the Roman Republican coins. Their algorithm uses a linear SVM on the spatial extensions of the standard bag-of-words (BoVWs) representation for image classification. To this end, we compare our results with the simple BoVWs representation~\cite{anwar2015ancient} and its variant with a rectangular tiling (RT)~\cite{anwar2013bag} of 2$\times$2 for empirically selected vocabulary sizes as shown in Table~\ref{tab:results}. Moreover, our precision is higher, which means the measurements are consistent. Similarly, higher recall indicates the results are relevant, returned by our network compared to competitive methods.

Our method performs better from the classical algorithms with an improvement of \textbf{27.7\%} and \textbf{10.1\%} on BoVWs~\cite{anwar2015ancient} and RT~\cite{anwar2013bag}, respectively. Furthermore, to compare with the current state-of-the-art convolutional neural networks \ie VGG~\cite{simonyan2014vgg} and NasNet~\cite{zoph2018learning}, we fine-tune the networks from imageNet~\cite{deng2009imagenet} using our coins' training set. The improvement is on VGG~\cite{simonyan2014vgg} and NasNet~\cite{zoph2018learning} is \textbf{1.1\%} and \textbf{0.7\%}, respectively. The improvement on CNN is less as compared to the traditional classifiers as the CNN methods may be benefiting from the weights of ImageNet~\cite{deng2009imagenet}.   

\subsection{Qualitative Comparison}

In Figure~\ref{fig:im_classify}, we show the correct and incorrect classification results on the randomly selected images from the original test dataset. The results are only furnished for the CNN algorithms' \ie VGG~\cite{simonyan2014vgg}, NasNet~\cite{zoph2018learning} and our CoinNet. 

In the top row of Figure~\ref{fig:im_classify}, our method, and NasNet~\cite{zoph2018learning} both can classify the input images correctly; hence marked with different shades of green circles and a confidence score at the top of each image. It can be observed that the confidence level of NasNet~\cite{zoph2018learning} is always lower even the prediction is correct compared to our CoinNet method. Likewise, we present the misclassification of the coin types by our method and VGG~\cite{simonyan2014vgg} in the bottom row of Figure~\ref{fig:im_classify} marked with red circles and again having the confidence score at the top of the image. In this case, VGG~\cite{simonyan2014vgg} is always more confident \ie, having a high score than our network. This sums up that our model is more confident about correct predictions and vice versa.  

\subsection{Network Visualization For Attention}

\begin{figure}
\begin{center}
\resizebox{\columnwidth}{!}{
\begin{tabular}{c@{}c@{ }c@{ }c@{ }c}
    \raisebox{1\normalbaselineskip}[0pt][0pt]{\rotatebox{90}{Dolphin}}&
    \includegraphics[width=0.25\linewidth, keepaspectratio]{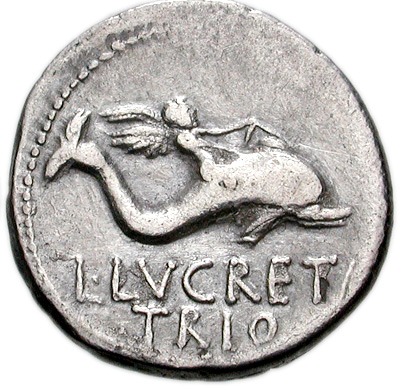}&  
    \includegraphics[width=0.25\linewidth, keepaspectratio]{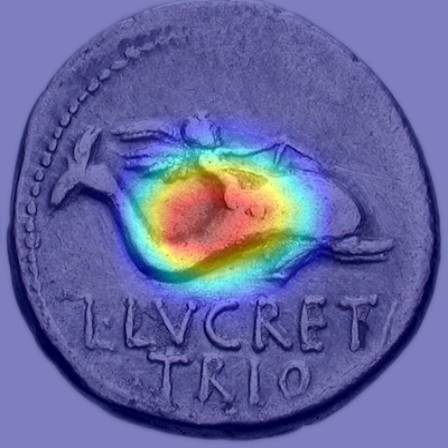}&  
    \includegraphics[width=0.25\linewidth, keepaspectratio]{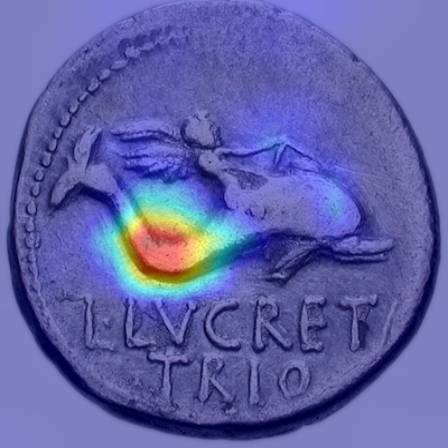}&  
    \includegraphics[width=0.25\linewidth, keepaspectratio]{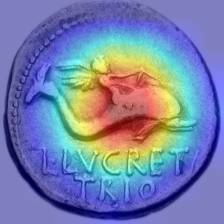}\\  
    
    \raisebox{1.5\normalbaselineskip}[0pt][0pt]{\rotatebox{90}{Biga}}&
    \includegraphics[width=0.25\linewidth, keepaspectratio]{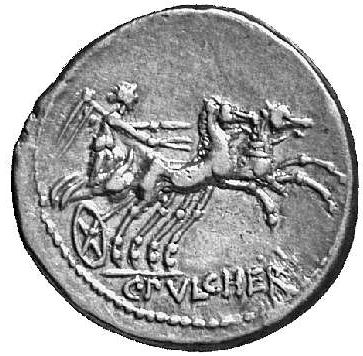}&  
    \includegraphics[width=0.25\linewidth, keepaspectratio]{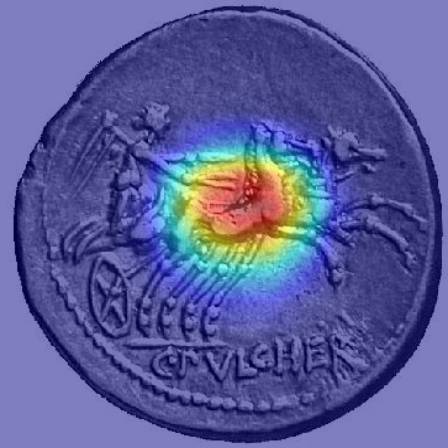}&  
    \includegraphics[width=0.25\linewidth, keepaspectratio]{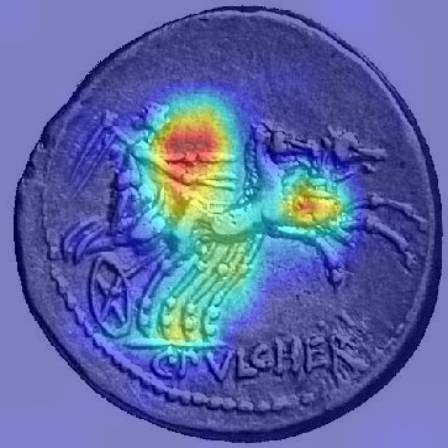}&  
    \includegraphics[width=0.25\linewidth, keepaspectratio]{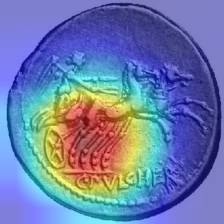}\\  
    
    \raisebox{1\normalbaselineskip}[0pt][0pt]{\rotatebox{90}{Minerva}}&
    \includegraphics[width=0.25\linewidth, keepaspectratio]{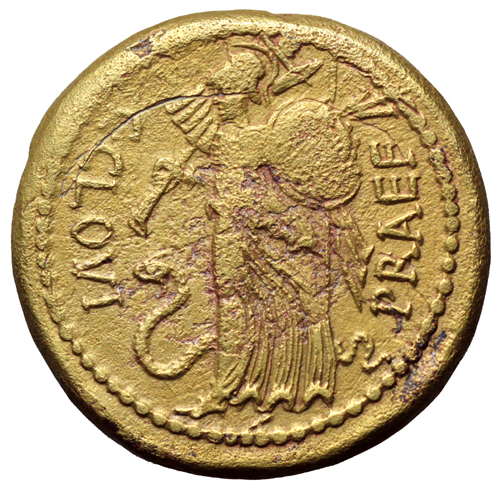}&  
    \includegraphics[width=0.25\linewidth, keepaspectratio]{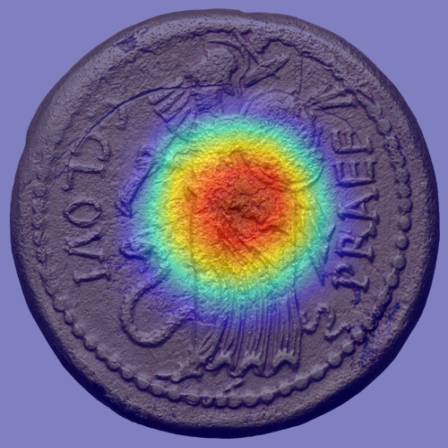}&  
    \includegraphics[width=0.25\linewidth, keepaspectratio]{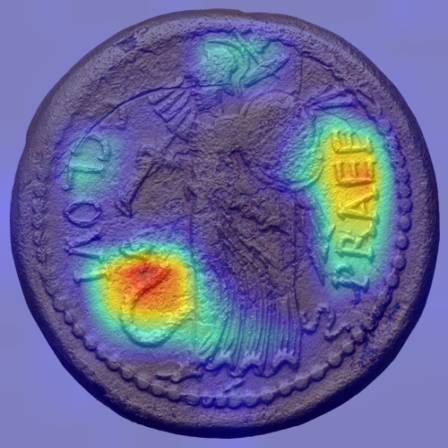}&  
    \includegraphics[width=0.25\linewidth, keepaspectratio]{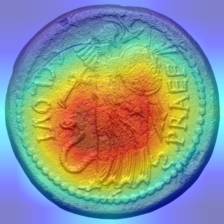}\\  
    
     & Input & VGG & NASNet &  CoinNet \\
 \end{tabular}
 }
\end{center}
\caption{\textbf{Visualization results from Grad-CAM~\cite{selvaraju2017grad}:} The visualization is computed for the last convolutional outputs, and the ground-truth labels are shown on the left column the input images.}
\label{fig:im_att}
\end{figure}

To visualize the importance of the feature attention, we employ a recently introduced method called Grad-CAM~\cite{selvaraju2017grad}. By computing the gradients concerning an individual class, Grad-CAM~\cite{selvaraju2017grad}, gives an insight into essential regions the network focuses. In Figure~\ref{fig:im_att} we provide a visualization comparison for VGG~\cite{simonyan2014vgg}, NasNet~\cite{zoph2018learning} and CoinNet.   

The first image in Figure~\ref{fig:im_att}, we can observe that the Grad-CAM~\cite{selvaraju2017grad} masks of our CoinNet network cover the \enquote{dolphin} object regions better than other methods where VGG~\cite{simonyan2014vgg} only focuses on a subpart of the object while NasNet~\cite{simonyan2014vgg} aims for non-essential regions. Similarly, in the \enquote{Biga} image, our method focuses on the number of horse legs while other CNN networks conform to either human head or horse abdomen, which can be found in the different coin images as well; hence resulting in low accuracy. As the last example, we present attention on \enquote{Minerva} coin image. As usual, VGG~\cite{simonyan2014vgg} focus on the middle part of the coin while NasNet~\cite{zoph2018learning} aim for text regions; however, our CoinNet model learns from more holistic feature regions as shown in the last row of Figure~\ref{fig:im_att}. The mentioned examples show that our CoinNet exploits and learns the essential information in the target objects and aggregate features for classifications from these regions, which helps in increasing accuracy.  
\subsection{Influence of Feature Maps}

We test the robustness of our network to the input image embeddings required for classification of the coins. For this purpose, we utilized the combinations of VGG~\cite{simonyan2014vgg}, ResNet~\cite{he2016deep} and DenseNet~\cite{huang2017densely}. Table~\ref{tab:Ablation_CNN} shows that the classification rate has a marginal difference as we employ another input embeddings. The leading cause for this phenomenon is that the network is not mainly relying on the input embeddings as MCB, residual blocks, and attention plays the primary role in learning the subtle variations among the coins.

\begin{table}
\caption{\textbf{Input features effect:} Comparison of different input features combinations to our CoinNet. Our network is robust to the change in the input features such as generated via ResNet50 (r50), DenseNet161 (d161) and Vgg19.}
\resizebox{0.8\columnwidth}{!}{
\centering
\begin{tabular}{|l|c|c|c|c|c|c|}
\hline
Nets  &r50-r50      &d161-d161  &r50-d161   & vgg19-r50 & vgg19-d161  \\ \hline
Acc.  & 98.4\%      & 98.5\%    & 98.5\%    & 98.4\%    & 98.5\%  \\ \hline
\end{tabular}
}
\label{tab:Ablation_CNN}
\vspace*{5mm}
\centering
\caption{\textbf{Influence of vocabulary:} The effect of the vocabulary size on the classification performance for BoVWs and rectangular tiling.}
\begin{tabular}{|c|c|c|}
\hline
\textbf{Vocabulary size}& \multicolumn{1}{l|}{\textbf{BoVWs}} & \multicolumn{1}{l|}{\textbf{RT}} \\ \hline
1k  & 65.80\%  & 84.44\%   \\ \hline
5k  & 70.81\%  & 83.80\%   \\ \hline
10k & 69.45\%  & 81.53\%   \\ \hline
15k & 69.15\%  & 79.81\%   \\ \hline
\end{tabular}
\label{tab:as}
\end{table}

We also perform ablation studies to get the best vocabulary size of the BoVWs representation where the vocabulary sizes are empirically selected, as shown in Table~\ref{tab:as}. An overfitting effect can be observed with an increase in vocabulary size. This effect is more noticeable in rectangular tiling, where the feature vector size that represents the image is four times the vocabulary size.

\subsection{Generalization Capability}
\label{subsec:transfer}
Here, we assess the generalization capability of our CoinNet. To this end, the model is trained with images of the 100 classes included in the original dataset. The models are then tested using the photos of the disjointed test set. Since the test images are disjoint, and there is no class label for the disjoint test images in the original dataset, we use a workaround where a test image of the object \enquote{Biga} will be considered as correctly classified if and only if it falls into any one of those 12 coin classes with \enquote{Biga} as the main object.  

Table~\ref{tab:BIGA_classify} presents the quantitative results where our CoinNet leads the other competitive state-of-the-art methods with a significant margin of more than $30\%$, thus demonstrating a far superior generalization performance of CoinNet on disjoint coin types. The performance increase can be partially attributed to the ResNet blocks, followed by the attention mechanism. 

\begin{table}
\caption{\textbf{Performance on disjoint set:} Accuracy on the unseen coin types for competing CNNs}
\centering
\begin{tabular}{|l|c|c|c|}
\hline
 & \multicolumn{3}{c|}{Methods} \\ \cline{2-4}
Datasets & VGG     & NASNet   & Ours               \\ \hline \hline
Biga       & 69.15\% & 48.64\% & \textbf{96.56}\% \\ 
Quadriga   &  4.37\% & 16.33\% & \textbf{68.15}\% \\ 
Curule     & 71.17\% & 8.11\%  & \textbf{79.28}\% \\ \hline \hline
\end{tabular}
\label{tab:BIGA_classify}
\end{table}

\subsection{Limitations}
\label{subsec:limitation}
Although the performance of CoinNet has surpassed the classical and CNN methods; however, like competitive methods, it still struggles to recognize the objects in the images due to the lower resolution. Few examples are previously presented in the second row of Figure~\ref{fig:im_classify}, where the images are either low-resolution or having blur in them; hence, it results in misclassification.

\section{Conclusion}
\label{sec:conclusion}
The classification of ancient Roman Republican coins via recognizing objects on their reverse sides is performed on a  new dataset comprised of diverse coin images. Our method outperformed the traditional BoVWs model and its spatial extensions that previously gave state-of-the-art results on the task of ancient coins classification. It was experimentally shown that on a large scale image dataset, the BoVWs model performs inferior and tends to overfit. We also compared our proposed CoinNet architecture with the current state-of-the-art CNN model, which lags in accuracy. Besides, our CoinNet also outperformed the competing CNNs on the unseen disjoint test set. In the future, we plan to recognize other visual cues of the reverse motifs that will ultimately support the current classification system for a more detailed classification of the coin.  

\bibliographystyle{elsarticle-num} 
\bibliography{refs}

\end{document}